\documentclass[10pt,journal,compsoc]{IEEEtran}

\ifCLASSOPTIONcompsoc
  \usepackage[nocompress]{cite}
\else
  \usepackage{cite}
\fi
\ifCLASSINFOpdf
\else
\fi

\usepackage{graphicx}
\usepackage{booktabs} 
\usepackage{adjustbox}
\usepackage{subcaption}
\usepackage{multirow}

\usepackage{algorithm}

\usepackage{amsmath}
\usepackage{amssymb}
\usepackage{xcolor}
\usepackage{breqn,xspace}
\usepackage{bm}
\usepackage{lipsum,multicol}
\usepackage{makecell}
\usepackage{booktabs}
\usepackage{algorithm}
\usepackage{amsmath}

\usepackage{algpseudocode}
\usepackage{amsmath}
\usepackage{diagbox}
\allowdisplaybreaks[2]

\ifodd 0
\definecolor{darkgreen}{RGB}{0,200,0}
\newcommand{\rev}[1]{{\color{blue}#1}} 
\else
\newcommand{\rev}[1]{#1}
\fi

\begin{document}
%
\title{\huge{Efficient Parallel Split Learning over Resource-constrained Wireless Edge Networks}}
%
%
%
%

\author{Zheng~Lin, Guangyu~Zhu, Yiqin~Deng,~\IEEEmembership{Member,~IEEE,} Xianhao Chen,~\IEEEmembership{Member,~IEEE}, \\ Yue Gao,~\IEEEmembership{Senior Member,~IEEE}, Kaibin Huang,~\IEEEmembership{Fellow,~IEEE},  and Yuguang Fang,~\IEEEmembership{Fellow,~IEEE}

\IEEEcompsocitemizethanks{\IEEEcompsocthanksitem
Zheng Lin, Xianhao Chen, and Kaibin Huang are with the Department of Electrical and Electronic Engineering, University of Hong Kong, Pok Fu Lam,
Hong Kong SAR, China. Zheng Lin was with the School of Computer Science, Fudan University, Shanghai, 200438, China. Xianhao Chen is also with HKU Musketeers Foundation Institute of Data Science, University of Hong Kong, Pok Fu Lam, Hong Kong SAR, China (e-mail: linzheng@eee.hku.hk; xchen@eee.hku.hk; huangkb@eee.hku.hk).
\IEEEcompsocthanksitem
Guangyu Zhu is with the Department of Electrical and Computer Engineering, University of Florida, Gainesville, FL 32611 USA (e-mail: gzhu@ufl.edu).
\IEEEcompsocthanksitem
Yiqin Deng is with the School of Control Science and Engineering, Shandong University, Jinan 250061, Shandong, China (e-mail: yiqin.deng@email.sdu.edu.cn).
\IEEEcompsocthanksitem
Yue Gao is with the School of Computer Science, Fudan University, Shanghai 200438, China (email: gao\underline{~}yue@fudan.edu.cn).
\IEEEcompsocthanksitem
Yuguang Fang is with the Department of Computer Science, City University of Hong Kong, Kowloon, Hong Kong SAR, China (e-mail: my.fang@cityu.edu.hk).
}

\thanks{\textit{\quad\quad\!\!(Corresponding author: Xianhao Chen)}}}

%
%

\markboth{Journal of \LaTeX\ Class Files,~Vol.~14, No.~8, August~2015}%
{Shell \MakeLowercase{\textit{et al.}}: Bare Advanced Demo of IEEEtran.cls for IEEE Computer Society Journals}
%



\IEEEtitleabstractindextext{%
\begin{abstract}
The increasingly deeper neural networks hinder the democratization of privacy-enhancing distributed learning, such as federated learning (FL), to resource-constrained devices. To overcome this challenge, in this paper, we advocate the integration of edge computing paradigm and parallel split learning (PSL), allowing multiple edge devices to offload substantial training workloads to an edge server via layer-wise model split. By observing that existing PSL schemes incur excessive training latency and large volume of data transmissions, we propose an innovative PSL framework, namely, efficient parallel split learning (EPSL), to accelerate model training. To be specific, EPSL parallelizes client-side model training and \textit{reduces the dimension of activations' gradients} for back propagation (BP) via \textit{last-layer gradient aggregation}, leading to a significant reduction in server-side training and communication latency. Moreover, by considering the heterogeneous channel conditions and computing capabilities at edge devices, we jointly optimize subchannel allocation, power control, and cut layer selection to minimize the per-round latency. Simulation results show that the proposed EPSL framework significantly decreases the training latency needed to achieve a target accuracy compared with the state-of-the-art benchmarks, and the tailored resource management and layer split strategy can considerably reduce latency than the counterpart without optimization.
\end{abstract}

}

\maketitle

\IEEEdisplaynontitleabstractindextext

%
\IEEEpeerreviewmaketitle


\section{Introduction}
With the proliferation of Internet of Things (IoT) devices and advancement of information and communications technology (ICT), networked IoT devices are capable of collecting massive data in a timely fashion. It is predicted in~\cite{report} that the worldwide number of IoT devices is expected to reach 29 billion by 2030, nearly tripling from 2020. The unprecedented amount of data generated from countless devices serves as the fuel to power artificial intelligence (AI), which has gained tremendous success in major sectors, including smart healthcare, natural language processing, and intelligent transportation~\cite{2022Letaief,zhu2020toward,hu2023adaptive,wu2020fedhome,hu2024collaborative,lin2022channel,wei2023differential}.

\begin{figure*}[t!]
\centering
\includegraphics[width=15.2 cm]{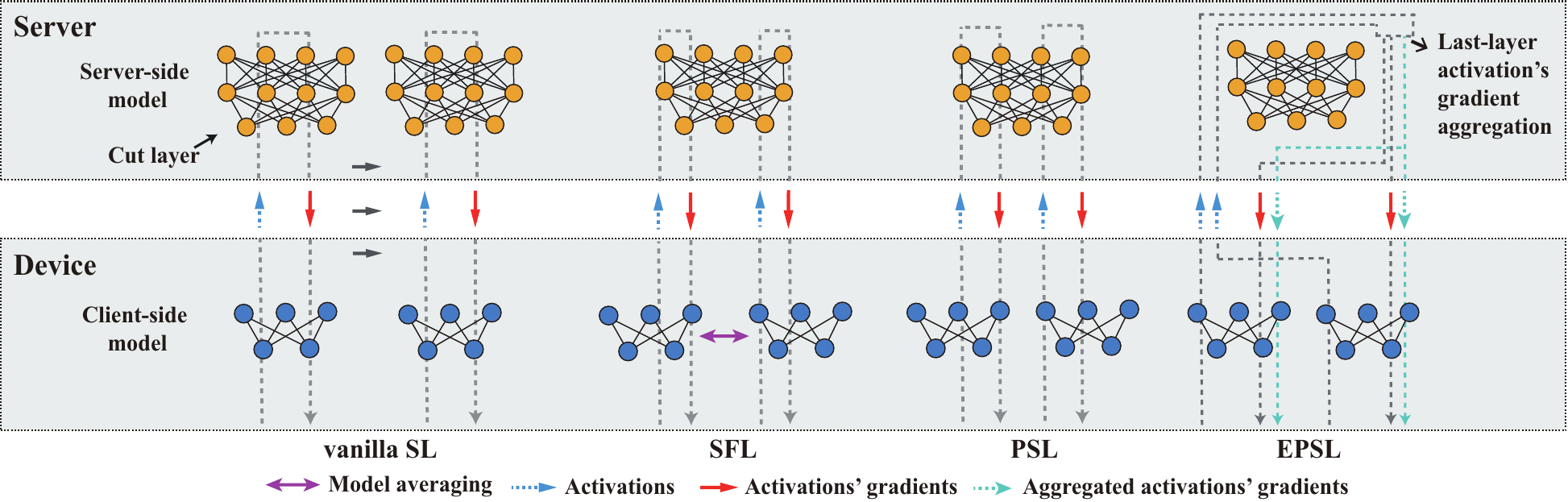}
\vspace{-0cm}
\caption{An illustration of vanilla SL, PSL, SFL and EPSL frameworks, where $\phi$ denotes the ratio of last-layer gradient aggregation. When $\phi = 0$, EPSL is reduced to PSL.
}
\label{DCML_comparison}
\end{figure*}

The conventional machine learning (ML) framework, known as centralized learning (CL), entails gathering and processing the raw data in a central server. Each edge device shares local raw data with a central server for model training. However, the lack of trust and privacy regulations deter edge devices from sharing their private data, hindering scalable AI deployment. Moreover, it is often unrealistic for massive edge devices to transfer local raw data to the remote cloud due to the large bandwidth consumption. Motivated by this, federated learning (FL)~\cite{konevcny2016federated} is proposed to enable collaborative model training without sharing raw data. Nevertheless, FL exclusively employs on-device training, which places heavy computation burden on edge devices. For instance, the popular image recognition model VGG16~\cite{VGG} comprises 138 million parameters (528MB in 32bit float) and requires 15.5G MACs (multiply-add computation) for forward propagation (FP) process, essentially making exclusive on-device training impractical for resource-constrained mobile or IoT devices. One promising solution is to compress the ML model to lower both communication and computing workload~\cite{shi2022toward,chen2022energy,xu2022adaptive}. However, model compression/quantization requires specialized hardware to take full advantage of low-bit computation and induces learning errors due to the low precisions.

To address the above challenges, split learning (SL)~\cite{vepakomma2018split} has emerged as an effective approach to taking over training load from edge devices while still preserving their raw data locally~\cite{lin2023split,lyu2023optimal}. Specifically, SL allows edge devices to offload substantial training workloads to a server via layer-wise model partitioning~\cite{lin2023pushing}. In the vanilla SL, the sequential training from one edge device to another, however, incurs excessive training latency. To overcome this severe limitation, split federated learning (SFL)~\cite{thapa2022splitfed} and parallel split learning (PSL) (without client-side model synchronization compared with SFL)~\cite{kim2022bargaining,joshi2021splitfed} have been proposed, which enable multiple devices to train in parallel, thus positioning SL as a promising alternative for FL in many situations.

Unfortunately, the existing SFL/PSL frameworks can still lead to significant computing and communication latency. On the one hand, an SL/SFL/PSL server \textit{takes over the main training workloads from multiple clients}. Although training models for a reasonable number of edge devices might not be a crucial issue for a sufficiently powerful cloud computing center, it is arguably overwhelming to a resource-constrained edge server as the number of served edge devices increases. Particularly, edge computing servers in 5G and beyond can be (small) base stations and access points usually equipped with limited capabilities~\cite{ding2018beef,deng2022actions,chen2022multi,fang2022age}. On the other hand, communication latency is a limiting factor due to the large volume of cut-layer data and model exchange involved in SL. Specifically, PSL incurs large volumes of cut-layer data exchange while SFL involves both massive cut-layer data exchange and model exchange, as summarized in Fig.~\ref{DCML_comparison}. To deploy low-latency SL at the network edge, \textit{a more efficient SL is desired}.

In this paper, we propose a novel SL framework, namely, efficient parallel split learning (EPSL), which is particularly useful for resource-constrained edge computing systems. The proposed EPSL framework incorporates parallel model training and model partitioning, which amalgamates the advantages of FL and SL.  The parallel training of client-side model efficiently utilizes idle local computing resources, while model partitioning significantly lowers the computation workload and communication overhead on edge devices. More importantly, we design the last-layer gradient aggregation on the server-side model to shrink the dimension of activations' gradients and computation workload on a server, leading to a considerable reduction in training latency compared with the state-of-the-art SL benchmarks, including SFL and PSL. Furthermore, this operation also reduces communication overhead and eliminates the necessity for model exchange. We define aggregation ratio $\phi  \in \left[ {0,1} \right] $  to control the fraction of last-layer activations' gradients to be aggregated, which strikes the balance between learning accuracy and latency reduction. The detailed comparison of FL, vanilla SL, SFL, PSL and EPSL frameworks is illustrated in Table 1, demonstrating that EPSL is more compute- and communication-efficient than existing SL schemes. It is noted that PSL is a special case of EPSL when the aggregation ratio is set to 0.

Moreover, at the wireless edge, the heterogeneous channel conditions and computing capabilities at edge devices considerably decrease the efficiency of ML model training due to the straggler effect~\cite{chen2021distributed,xiao2023time,lee2017speeding}. In this regard, the appropriate design of resource management and layer split strategy can preferably allocate more resources to the straggler, thereby boosting learning efficiency. Towards this end, we formulate an optimization problem of subchannel allocation, power control, and cut layer selection for the proposed EPSL, and develop an efficient algorithm to solve it. The major contributions of this paper are summarized as follows.
\begin{table}[t]\label{table1}
\caption{The Comparison of FL, vanilla SL, SFL, PSL and EPSL Frameworks.}
  \renewcommand{\arraystretch}{1.4}{
  \setlength{\tabcolsep}{2mm}{
\begin{tabular}{cccccc}
\hline
\textbf{Learning framework}                                                              & \textbf{FL} & \textbf{vanilla SL} & \textbf{SFL} & \textbf{PSL} & \textbf{EPSL} \\ \hline
\textbf{\begin{tabular}[c]{@{}c@{}}Partial computation\\ offloading  \end{tabular}} & No          & Yes     & Yes    & Yes          & Yes           \\ \hline
\textbf{\begin{tabular}[c]{@{}c@{}}Parallel computing\end{tabular}}           & Yes    & No  & Yes  & Yes   & Yes      \\ \hline
\textbf{Model exchange}     & Yes & No & Yes & No & No  \\ \hline
\textbf{\begin{tabular}[c]{@{}c@{}}Activations' gradients'\\dimension reduction\end{tabular}} & No & No & No & No & Yes \\ \hline
\textbf{Access to raw data}    & No  & No  & No  & No  & No\\ \hline
\end{tabular}}}
\end{table}
\vspace{1em}
\begin{itemize}

\vspace{-0.15cm}
\item  We propose EPSL, a compute- and communication-efficient PSL framework, which features the key operation of aggregating the last-layer activations' gradients during the BP process to reduce communication and computing overhead for PSL.
\item  We design an effective resource management and layer split strategy that jointly optimizes subchannel allocation, power control, and cut layer selection for EPSL to minimize the per-round latency.
\item  We conduct extensive simulation studies to demonstrate that the proposed EPSL framework significantly decreases the training latency needed to achieve a target accuracy compared with existing SL schemes. Besides, the tailored resource management and layer split strategy can significantly reduce the latency than the counterpart without optimization.
\end{itemize}

The remainder of this paper is organized as follows. Section~\ref{Related_Work} and~\ref{System_Model} introduce the related work and system model. Section~\ref{EPSL_Framework} elaborates on the detailed EPSL framework. We formulate the resource management problem in Section~\ref{PROBLEM_FORMULATION} and offer the corresponding solution approach in Section~\ref{Solution_Approach}. Section~\ref{simulation results} covers the simulation results. Finally, concluding remarks are presented in Section~\ref{conclusion_section}.


\section{Related Work}\label{Related_Work}

FL has become the most prevalent distributed learning framework due to its advantages in data privacy and parallel model training. However, since FL is resource-hungry, the limited communication and computing capabilities at edge devices are the bottlenecks~\cite{shi2022toward}. \rev{Tremendous research efforts have been made to optimize wireless edge networks for effective FL deployment, covering the aspects like client selection~\cite{xu2020client,pang2022incentive}, resource allocation~\cite{jiao2020toward,yang2020energy,chen2020joint,dinh2022network}, hierarchical model aggregation~\cite{liu2020client,chen2022federated,liu2022time,lin2023fedsn}, and model compression~\cite{shi2022toward,chen2022energy,chen2023service}. }

In contrast to FL, SL has recently emerged as a novel privacy-enhancing distributed learning framework mainly targeting resource-constrained devices. It was first proposed by Vepakomma~\textit{et al.}~\cite{vepakomma2018split} for health systems in which privacy concern is of high priority. To enable privacy-preserving deep neural network computation, Kim~\textit{et al.}~\cite{kim2021spatio} develop a novel SL framework with multiple end-systems and verify by experiments that the proposed framework achieves near-optimal accuracy while guaranteeing data privacy. In~\cite{yang2022differentially}, a generic gradient perturbation-based split learning framework is designed to provide provable differential privacy guarantee. The earlier works focus on vanilla SL operating in a sequential fashion, resulting in excessive training latency. Thepa~\textit{et al.}~\cite{thapa2022splitfed} integrate FL and SL to parallelize SL to accelerate model training. Pal~\textit{et al.}~\cite{pal2021server} propose a more scalable PSL framework to address server-side large batch and the backward client decoupling problems by averaging the local gradients at the cut layer. However, our studies show that the training latency in SL can still be substantially reduced without noticeably impacting learning accuracy.

Apart from learning framework development, we will also investigate the resource management and layer split strategy tailored for the proposed SL frameworks. Along this line, Kim~\textit{et al.}~\cite{kim2022bargaining} present the optimal cut layer selection strategy for balancing the energy consumption, training time, and data privacy. To minimize the expected energy and time cost, Yan, Bi and Zhang~\cite{yan2022optimal} devise the joint optimization of model placement and model splitting for split inference. In~\cite{tian2022jmsnas}, a joint model split and neural architecture search approach is proposed to automatically generate and place neural networks over chain or mesh networks. Unfortunately, research on resource management strategy for enhancing SL performance is still in its infancy. It is obvious that resource management and model split are tightly coupled. Reference~\cite{wu2022split} is one of the few studies that considers the integrated design of resource management and cut layer selection, where device clustering is also exploited to reduce communication overhead. However, the algorithm implementation is hampered by the exhaustive search with high computational complexity.

\section{System Model}\label{System_Model}
\begin{figure}[t!]
\centering
\includegraphics[width=7.2cm]{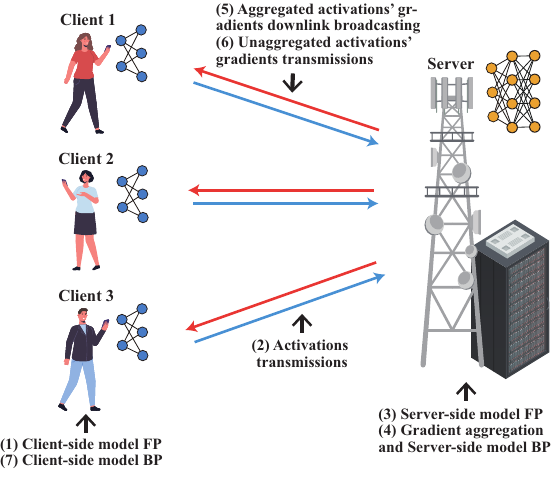}
\vspace{-0.2cm}
\caption{The illustration of EPSL over wireless networks.
}
\label{EPSL}
\end{figure}

As demonstrated in Fig.~\ref{EPSL}, we consider a typical scenario of EPSL framework in wireless networks, which consists of two essential components:

\begin{itemize}
\item \textbf{Edge devices:} We assume that each client  possesses an end device with computing capability to perform FP and BP processes for client-side model. The set of participating edge devices in the model training is denoted by $\mathcal{C} = \left\{ {1,2,...C} \right\}$, where $C$ is the number of edge devices. The local dataset ${\mathcal{D}_i}$ with ${D_i}$ data samples owned by edge device $i$ is represented as ${\mathcal{D}_i} = \left\{ {{{\bf{x}}_{i,k}},{y_{i,k}}} \right\}_{k = 1}^{{D_i}}$, where  ${{\bf{x}}_{i,k}} \in {\mathbb{R}^{d \times 1}}$ denotes the $k$-th input data in the local dataset ${\mathcal{D}_i}$ and ${{{y}}_{i,k}} \in {\mathbb{R}^{1}}$ is the label of ${{\bf{x}}_{i,k}}$. Therefore, the total dataset ${\mathcal{D}}$ with $D = \sum\limits_{i = 1}^C {{D_i}} $ data samples is denoted by ${\cal D} =  \cup _{i = 1}^C{{\cal D}_i}$. The client-side model is represented as ${{\bf{W}}_c} \in {\mathbb{R}^{u \times 1}}$, where $u$ denotes the dimension of client-side model's parameters.
\item \textbf{Server:} The server is a central entity with powerful computing capability, which takes charge of server-side model training. The server-side model is denoted by ${{\bf{W}}_s} \in {\mathbb{R}^{r \times 1}}$, where $r$ represents the dimension of server-side model's parameter. The server gathers important network information such as channel status information and device computing capability, to implement the resource management strategy.
\end{itemize}

The global model is represented as ${\bf{W}} = \left[ {{{\bf{W}}_s};{{\bf{W}}_c}} \right] \in {\mathbb{R}^{\left( {r + u} \right) \times 1}}$. For edge device $i$, we assume that the activation obtained according to ${{{\bf{x}}_{i,k}}}$ is denoted by ${{\bf{s}}_{i,k}} = f\left( {{{\bf{x}}_{i,k}};{{\bf{W}}_c}} \right)\in {\mathbb{R}^{q \times 1}}$, where $q$ denotes activation's dimension per data sample, $f\left( {{\bf{x}};{\bf{w}}} \right)$ represents the mapping function reflecting the relationship between input data ${\bf{x}}$ and its predicted value given model parameter ${\bf{w}}$. Likewise, based on the activation ${{{\bf{s}}_{i,k}}}$, we denote the predicted value as ${{\hat y}_{i,k}} = f\left( {{{\bf{s}}_{i,k}};{{\bf{W}}_s}} \right)$. Therefore, the local loss function for edge device $i$ is ${L_i}\left( {\bf{W}} \right) = \frac{1}{{ {{{D}_i}} }}\sum\limits_{k = 1}^{ {{{D}_i}} } {{L_{i,k}}\left( {{{\hat y}_{i,k}},{y_{i,k}}};{\bf{W}} \right)} $, where ${{L_{i,k}}\left( {{{\hat y}_{i,k}},{y_{i,k}};{\bf{W}}} \right)}$ represents the sample-wise loss function of the predicted value and the label for the $k$-th data sample in the local dataset ${\mathcal{D}_i}$. The global loss function $L\left( {\bf{W}} \right)$ is the weighted average of all participating edge devices' local loss functions. The objective of SL is to seek the optimal model parameter ${{\bf{W}}^{\bf{*}}}$ for the following optimization problem to minimize the global loss function:
\begin{align}\label{minimiaze_loss_function}
\mathop {\min }\limits_{\bf{W}} L\left( {\bf{W}} \right) = \mathop {\min }\limits_{\bf{W}} \sum\limits_{i = 1}^C {\frac{{{D_i}}}{D}} {L_i}({\bf{W}}).
\end{align}
To solve~\eqref{minimiaze_loss_function}, the conventional SL employs the sequential model training to find the optimal model parameters, resulting in a dramatic increase in training latency. In view of this, we propose a state-of-the-art EPSL framework, which incorporates parallel model training and model partitioning. For readers' convenience, the important notations in this paper are summarized in Table 2.

\begin{table}[t]\label{notation}
\caption{Summary of Important Notations.}
  \renewcommand{\arraystretch}{1.1}{
  \setlength{\tabcolsep}{1mm}{
\begin{tabular}{ll}
\hline
\textbf{Notation}                                                              & ~~~\textbf{Description}  \\ \hline
~~~$\mathcal{C}$ &  ~~~The set of participating edge devices     \\
~~~${\mathcal{D}_i}$           &  ~~~The local dataset of edge device $i$  \\
~~~$L\left( {\bf{W}} \right)$     &  ~~~The global loss function with model \\
&  ~~~parameter ${{\bf{W}}}$ \\
~~~$\mathcal{T}$     &  ~~~The set of training rounds \\
~~~${\mathcal{B}_i}$     &  ~~~The mini-batch drawn from client \\
&  ~~~device $i$'s local dataset\\
~~~$b$     &  ~~~The mini-batch size \\
~~~${{\bf{W}}_s}$           &  ~~~The server-side model  \\
~~~${{\bf{W}}_{c,i}}$           &  ~~~The client-side model at edge device $i$  \\
~~~$\ell_s$/$\ell_c$           &  ~~~The number of server-side/client-side\\
&  ~~~model layers  \\
~~~${{f _s}}$/${{f _i}}$           &  ~~~The computing capability of server/client\\
&  ~~~device $i$  \\
~~~${{\kappa _s}}$/${{\kappa _i}}$            &  ~~~The computing intensity of server/client\\
&  ~~~device $i$  \\
~~~${{B_k}}$            &  ~~~The bandwidth of the $k$-th subchannel  \\
~~~$G_s$/$G_c$           &  ~~~The effective antenna gain of the server/a \\
&  ~~~edge device  \\
~~~$\gamma ({F_k},{d_i})$            &  ~~~The average channel gain from edge device $i$ to\\
&  ~~~the server  \\
~~~${{p^{DL}}}$            &  ~~~The transmit PSD of the server  \\
~~~${{\sigma ^2}}$            &  ~~~The PSD of the noise  \\
~~~${\rho _j}$/${\varpi _j}$           &  ~~~The FP/BP computation workload of \\
& ~~~propagating the first $j$ layer neural \\
&  ~~~network for one data sample  \\
~~~${\psi _j}$/${\chi _j}$           &  ~~~The data size of the activations/activations'\\
&  ~~~gradients at the cut layer $j$\\
~~~$r_i^k$, $\mu_j$, $p_k$            &  ~~~Decision variables (explained in Section~\ref{PROBLEM_FORMULATION})  \\
~~~$T_1$, $T_2$           &  ~~~Auxiliary variables (to linearize problem~\eqref{sub2})  \\\hline
\end{tabular}}}
\end{table}

\section{EPSL Framework}\label{EPSL_Framework}
In this section, we present the proposed EPSL framework. The fundamental idea of EPSL is to parallelize the client-side model training while executing last-layer gradient aggregation to reduce the training latency. Besides computation workload and activations gradients' dimension reduction, the last-layer gradient aggregation eliminates the necessity for model exchange, which facilitates the implementation of EPSL.

Before model training begins, the server initializes the ML model and partitions it into a server-side model and a client-side model via layer-wise model partitioning (the cut layer selection strategy will be elaborated in Section~\ref{Solution_Approach}), then broadcasting the client-side model to all participating edge devices. After that,  EPSL is performed in consecutive training rounds until the model converges. As depicted in Fig.~\ref{EPSL}, for training round\footnote{{In each training round, we assume that every edge device employs a mini-batch with $b$ data samples from its local dataset for model training.}} $t \in \mathcal{T} = \left\{ {1,2,...,T} \right\}$, the EPSL training procedure consists of the following stages.

\begin{figure}[t!]
\centering
\includegraphics[width=6.5cm, height=6.0cm]{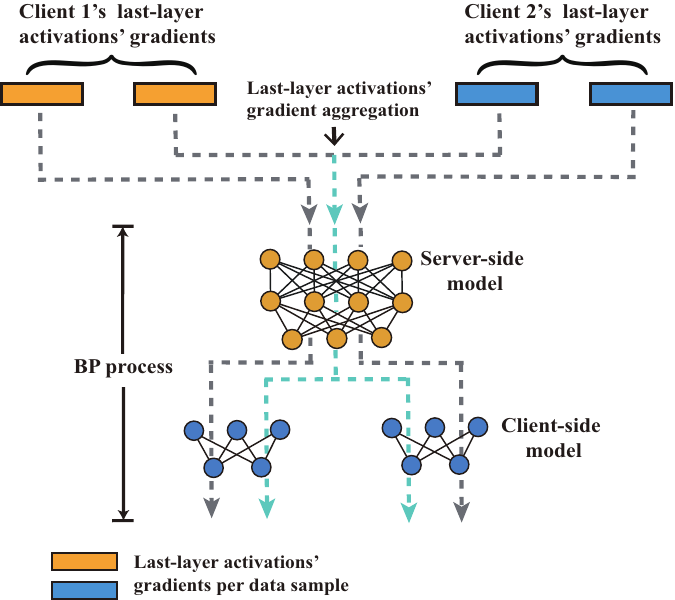}
\vspace{-0cm}
\caption{An example of the last-layer gradient aggregation in EPSL with $\phi=0.5$ (e.g., half of data going through last-layer aggregation), where two edge devices with two data samples participate in model training.
}
\label{last-layer_agg}
\end{figure}

\textit{1) Client-side Model Forward Propagation:}
At this stage, all participating edge devices perform the FP for client-side models in parallel. Specifically, for edge device $i$, a mini-batch ${\mathcal{B}_i} \subseteq {\mathcal{D}_i}$ with $b$ data samples is randomly drawn  from its local dataset. The input data and label of mini-batch are denoted by ${{\bf{X}}_i}\left( t \right) \in {\mathbb{R}^{b \times d}}$ and ${{\bf{y}}_i}\left( t \right) \in {\mathbb{R}^{b \times 1}}$, and ${\bf{W}}_{c,i}\left( t-1 \right)$ represents the client-side model of edge device $i$. After feeding the client-side model with the input data from the mini-batch, the cut layer generates the activations. The activations of edge device $i$ is given by
\begin{align}\label{stage_2}
{{\bf{S}}_i}\left( t \right) = f\left( {{{\bf{X}}_i}\left( t \right);{{\bf{W}}_{c,i}}\left( t-1 \right)} \right)\in {\mathbb{R}^{b \times q}}.
\end{align}

\textit{2) Activations Transmissions:}
After the FP process for the client-side model is completed, each participating edge device sends the activations and corresponding labels to the server over the wireless channel. The activations collected from all participating edge devices is utilized as the input to the server-side model.

\textit{3) Server-side Model Forward Propagation:}
After receiving the activations from participating edge devices, the server concatenates all activations to execute FP process for the server-side model. The concatenated activation matrix is denoted by ${\bf{S}}\left( t \right) = \left[ {{{\bf{S}}_1}\left( t \right);{{\bf{S}}_2}\left( t \right);...;{{\bf{S}}_C}\left( t \right)} \right] \in {\mathbb{R}^{bC \times q}}$, and the predicted value is represented as
\begin{align}\label{stage_4}
{\bf{\hat y}}\left( t \right) = f\left( {{\bf{S}}\left( t \right);{{\bf{W}}_s}\left( t-1 \right)} \right) \in {\mathbb{R}^{bC}}.
\end{align}

\textit{4) Gradient Aggregation and Server-side Model Back Propagation:}  At this stage, EPSL executes the last-layer gradient aggregation on the server-side model to shrink the dimension of {activations' gradients and computation workload of server, resulting in a considerable reduction in training latency. The server calculates the last-layer activations' gradients based on the loss function value, and then completes the server-side BP according to these gradients. The aggregation ratio $\phi  \in \left[ {0,1} \right] $  represents the fraction of activations' gradients that are aggregated at the last layer, i.e., $\left\lceil {\phi b} \right\rceil$ last-layer activations' gradients of every edge device are aggregated in a client-wise manner before executing BP process, and the remaining $\left( {b - \left\lceil {\phi b} \right\rceil } \right)$ unaggregated gradients are directly back-propagated. \rev{The aggregation ratio is a parameter that controls the trade-off between test accuracy and training latency. A higher aggregation ratio implies that more last-layer activations' gradients are aggregated, thereby reducing training latency at the cost of model accuracy. This is because a higher aggregation ratio implies that more personalized data features from edge devices are aggregated into shared features, resulting in the loss of information, whereas a lower aggregation ratio preserves more useful information.  In practice,  the aggregation ratio selection can be tailored for a specific training tasks by considering their model accuracy and latency requirements. For some non-critical tasks (e.g., virtual assistants),  higher aggregation ratio  can be selected to reduce the latency while ensuring a tolerable  compromise in accuracy, whereas for critical tasks (e.g., healthcare monitoring and diagnosis), a lower aggregation ratio  may be preferred.} One toy example is illustrated in Fig.~\ref{last-layer_agg}, where $\phi=0.5$ and hence half of back-propagated gradients are averaged before the BP process. It is worth noting that PSL is a special case of EPSL where $\phi = 0$. For edge device $i$, the client-side model is denoted by ${{\bf{W}}_{c,i}}\left( t \right) = \left[ {{\bf{w}}_{c,i}^{{\ell _c}}\left( t \right);{\bf{w}}_{c,i}^{{\ell _c} - 1}\left( t \right);...;{\bf{w}}_{c,i}^1\left( t \right)} \right]$, where ${\bf{w}}_{c,i}^k\left( t \right)$ is the $k$-th layer in the client-side model and $\ell _c$ represents the number of client-side model layers. The server-side model is represented as ${{\bf{W}}_s}\left( t \right) = \left[ {{\bf{w}}_s^{{\ell _s}}\left( t \right);{\bf{w}}_s^{{\ell _s} - 1}\left( t \right);...;{\bf{w}}_s^{ 1}\left( t \right)} \right]$, where ${\bf{w}}_s^{k}$ is the $k$-th layer in the server-side model and $\ell _s$ denotes the number of server-side model layers. Mathematically, the gradients of server-side model is expressed as
\begin{equation}\label{stage_5_1}
{{\bf{G}}_s}\left( t \right) = \left[ {\begin{array}{*{20}{c}}
{{\bf{g}}_s^{{\ell _s}}\left( t \right)}\\
{\!\!\!{\bf{g}}_s^{{\ell _s} - 1}\left( t \right)}\!\!\!\\
\vdots\\
{{\bf{g}}_s^1\left( t \right)}
\end{array}} \right],
\end{equation}
where ${\bf{g}}_s^{k}$ is the gradients of the server-side $k$-th layer, which is calculated by

\begin{equation}\label{gradient_k_layer}
{\bf{g}}_s^k \!=\! \!\!\left[ \!{\underbrace {\frac{1}{b},...,\frac{1}{b}}_{\left\lceil {\phi b} \right\rceil },\overbrace {\underbrace {\frac{{{\lambda _1}}}{b},...,\frac{{{\lambda _1}}}{b}}_{b - \left\lceil {\phi b} \right\rceil },...,\underbrace {\frac{{{\lambda _C}}}{b},...,\frac{{{\lambda _C}}}{b}}_{b - \left\lceil {\phi b} \right\rceil }}^{C\left( {b - \left\lceil {\phi b} \right\rceil } \right)}} \right]\!\!\!\left[ {\begin{array}{*{20}{c}}
{f\left( {\overline {{\bf{z}}_{s,1}^k} \left( t \right)} \right)}\\
 \vdots \\
{f\left( {\overline {{\bf{z}}_{s,{\left\lceil {\phi b} \right\rceil } }^k} \left( t \right)} \right)}\\
\!\!\!{f\left( {{\bf{z}}_{s,1,{\left\lceil {\phi b} \right\rceil }  + 1}^k\left( t \right)} \right)}\!\!\!\!\!\\
 \vdots \\
{f\left( {{\bf{z}}_{s,C,b}^k\left( t \right)} \right)}
\end{array}} \right]\!\!,
\end{equation}
 where ${\lambda _i} = \frac{{{D_i}}}{D}$, $f\left( \cdot \right)$ represents the function mapping activations' gradients to weights' gradients based on the standard BP process, $\overline {{\bf{z}}_{s,j}^k} \left( t \right)$ $\left( {j \in \left[ {1,\left\lceil {\phi b} \right\rceil} \right]} \right)$ denotes the $j$-th aggregated activations' gradients of server-side $k$-th layer, ${\bf{z}}_{s,i,{\left\lceil {\phi b} \right\rceil }  + j}^k\left( t \right)$ $\left( {j \in \left[ {1,b - \left\lceil {\phi b} \right\rceil } \right]} \right)$ represents the $j$-th unaggregated activations' gradients of server-side $k$-th layer for edge device $i$. According to the BP process,  $\overline{{\bf{z}}^{k}_{s,j}}(t)=\overline{{\bf{z}}^{k+1}_{s,j}}(t) \odot\Delta_{k}^{k+1}$, ${\bf{z}}^{k}_{s,i,{\left\lceil {\phi b} \right\rceil }+j}(t)={\bf{z}}^{k+1}_{s,i,{\left\lceil {\phi b} \right\rceil }+j}(t) \odot \Delta_{k}^{k+1}$, where $\odot$ denotes the element-wise product operation, $\Delta_{k}^{k+1}$ is the derivative of the $(k+1)$-th layer activations with respect to the $k$-th layer activations in the server-side model. Regarding the vectors on the right-hand side of~\eqref{gradient_k_layer}, the first $\left\lceil {\phi b} \right\rceil$ elements correspond to aggregated activations' gradients, and the remaining elements correspond to unaggregated activations' gradients.  \rev{It is clear that the dimension of the back-propagated gradients is reduced from the original size of $Cb$ to $\left\lceil {\phi b} \right\rceil  + C\left( {b - \left\lceil {\phi b} \right\rceil } \right)$, which significantly reduces the volume of exchanged data and training workload.} Obviously, to obtain ${\bf{G}}_s(t)$, the key is to construct $\overline{{{\bf{z}}_{s,j}^{{\ell _s}}}}\left( t \right)$ at the last layer and the subsequent steps are just standard BP. $\overline{{{\bf{z}}_{s,j}^{{\ell _s}}}}\left( t \right)$ is given by
\begin{equation}\label{last_aggregation}
\overline {{\bf{z}}_{s,j}^{{\ell _s}}} \left( t \right) = \sum\limits_{i = 1}^C {{\lambda _i}} {\bf{z}}_{s,i,j}^{{\ell _s}} {\kern 1pt}, \quad j \in \left[ {1,\left\lceil {\phi b} \right\rceil} \right]
\end{equation}
which means that the first $\lceil \phi b \rceil$ back-propagated gradients are averaged across the edge devices. Hence, the parameter update of the server-side model is expressed as
\begin{align}\label{stage_5_2}
{{\bf{W}}_s}\left( t \right) \leftarrow {{\bf{W}}_s}\left( t-1 \right) - {\eta}_s {{\bf{G}}_s}\left( t \right),
\end{align}
where ${\eta}_s$ is the learning rate for server-side model update.

\begin{figure}[t]
\centering
\begin{subfigure}{0.25\textwidth}
  \centering
  \includegraphics[width=4.3cm,height=3.7cm]{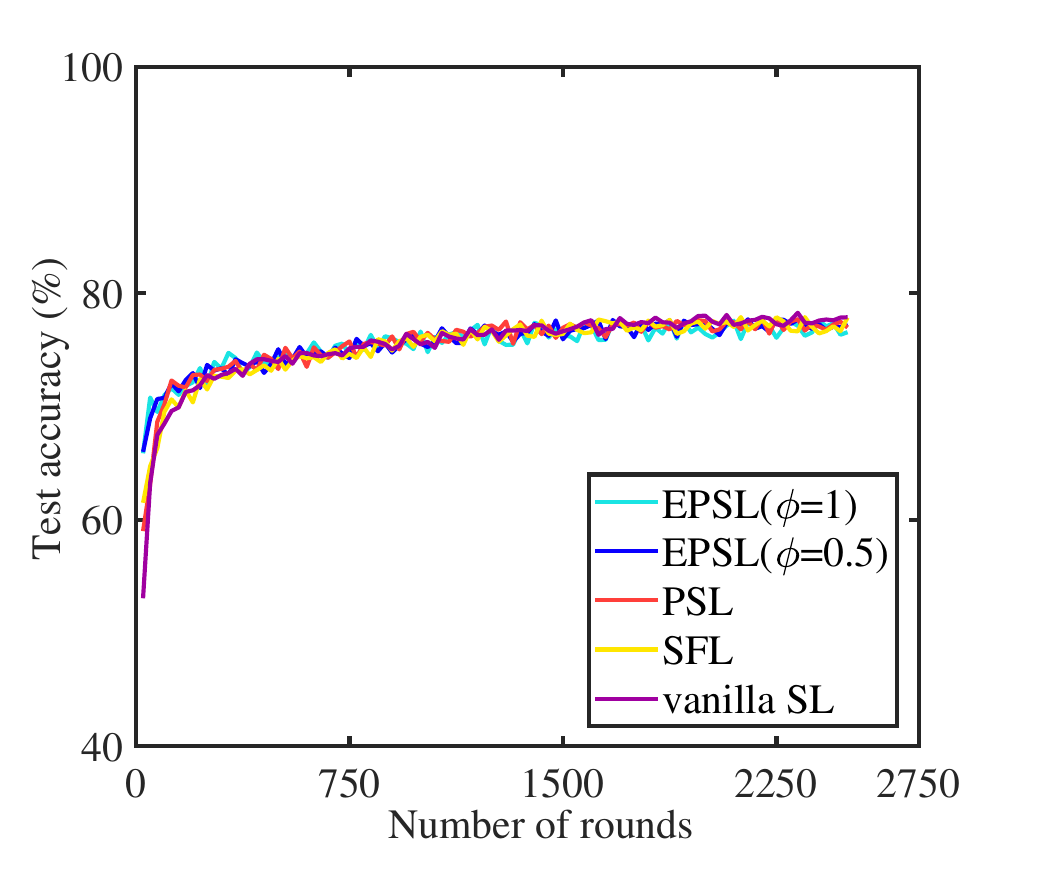}
  \caption{}
  \label{add_fig_1}
\end{subfigure}%
\begin{subfigure}{0.25\textwidth}
  \centering
  \includegraphics[width=4.4cm,height=3.7cm]{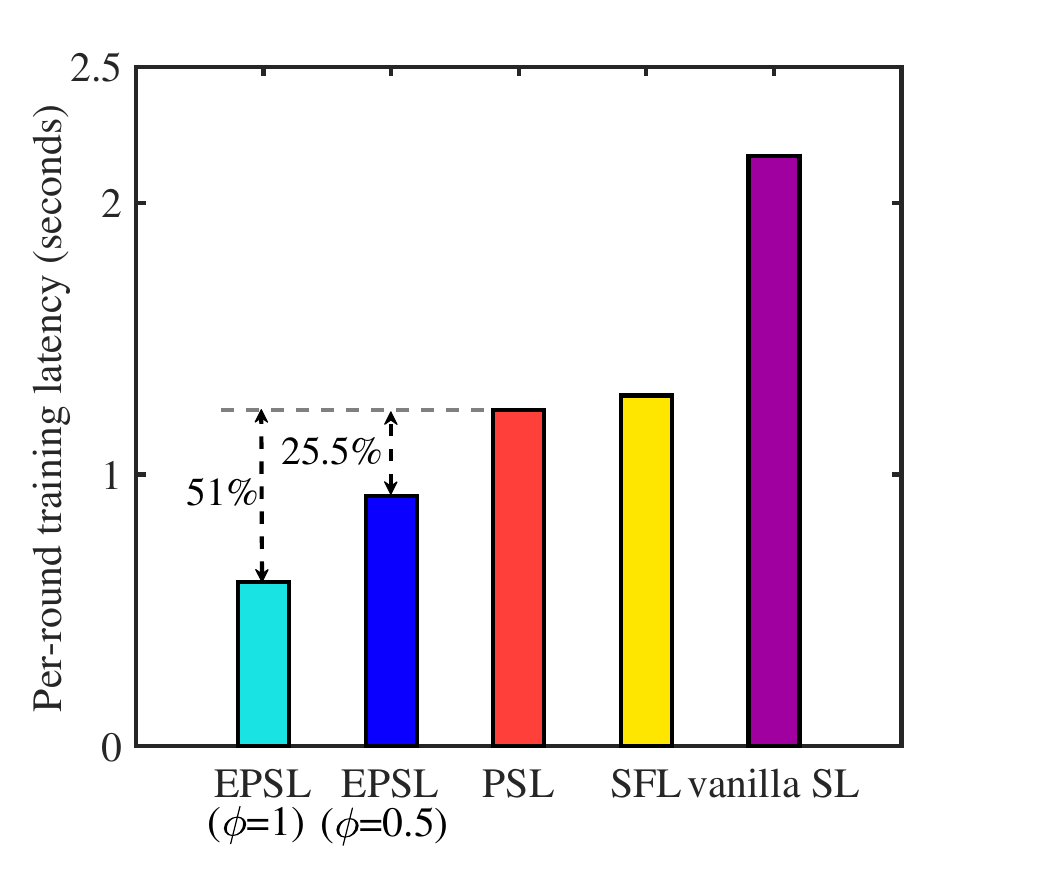}
  \caption{}
  \label{add_fig_2}
\end{subfigure}
\caption{The test accuracy (a) and per-round latency (b) of EPSL, PSL, SFL, and vanilla SL on HAM 10000 under IID settings using ResNet-18 with $C=5$ edge devices. As shown in the figures, EPSL achieves similar accuracy compared with other benchmarks with the same number of rounds, yet with much shorter per-round latency.}
\label{additional_fig}
\end{figure}

The last-layer gradient aggregation lies in the heart of EPSL, which remarkably decreases the communication and computing overhead in the BP process thereby boosting the efficiency in resource-constrained systems. Moreover, since the server broadcasts the aggregated gradients to all clients, the client-side models learn the shared features from each other, preventing overfitting without requiring model exchange as in FL or SFL. \rev{To better interpret EPSL, traditional frameworks (e.g., PSL) employ the conventional mini-batch stochastic gradient descent (SGD) for model training, i.e., finishing BP first and then averaging the gradients, while our approach averages the back-propagated gradients first and then conducts BP, thereby significantly reducing the times of BPs required. Due to the non-linearity of activation functions, these two are not equivalent. However, our approach might still be a good approximation as popular activation functions (e.g., ReLU and Sigmoid) are linear or approximately linear in a certain interval.} As shown in Fig.~\ref{additional_fig}, despite the significant per-round latency reduction, we find that EPSL with $\phi=0.5$ and $\phi=1$ achieve a similar learning accuracy versus the number of rounds as compared with vanilla SL, SFL and PSL.


\textit{5) Aggregated Activations' Gradients Downlink Broadcasting:}
When the BP process reaches the cut layer, the server broadcasts the aggregated activations' gradients (namely, the gradients that are generated at the cut layer by performing BP on the last-layer aggregated activations' gradients) to all participating edge devices over the shared wireless channel.

\textit{6) Unaggregated Activations' Gradients Transmissions:}
After the aggregated activations' gradients downlink broadcasting is completed, the server sends the unaggregated activations' gradients (i.e., the gradients that are obtained at the cut layer by executing BP on the last-layer unaggregated activations' gradients.) to corresponding edge devices over the wireless channel.

\textit{7) Client-side Model Back Propagation:}
At this stage, each edge device completes the parameter update of its client-side model based on the received activations' gradients. The client-side gradients for edge device $i$ can be calculated
\begin{align}\label{stage_7_1}
{{\bf{G}}_{c,i}}\left( t \right) = \left[ {\begin{array}{*{20}{c}}
{{\bf{g}}_{c,i}^{{\ell _c}}\left( t \right)}\\
{{\bf{g}}_{c,i}^{{\ell _c} - 1}\left( t \right)}\\
{...}\\
{{\bf{g}}_{c,i}^1\left( t \right)}
\end{array}} \right],
\end{align}
where ${\bf{g}}_{c,i}^{k}$ represents the gradients of the client-side $k$-th layer of edge device $i$, which is given by
\begin{align}\label{client_k_gradient}
{\bf{g}}_{c,i}^k = \Bigg[ {\underbrace {\frac{1}{b},...,\frac{1}{b}}_b} \Bigg]\left[ {\begin{array}{*{20}{c}}
{f\left( {\overline {{\bf{z}}_{c,1}^k} \left( t \right)} \right)}\\
 \vdots \\
{f\left( {\overline {{\bf{z}}_{c,{\left\lceil {\phi b} \right\rceil } }^k} \left( t \right)} \right)}\\
{f\left( {{\bf{z}}_{c,i,{\left\lceil {\phi b} \right\rceil }  + 1}^k\left( t \right)} \right)}\\
 \vdots \\
{f\left( {{\bf{z}}_{c,i,b}^k\left( t \right)} \right)}
\end{array}} \right],
\end{align}
where $\overline {{\bf{z}}_{c,j}^k} \left( t \right) = \overline {{\bf{z}}_{c,j}^{k + 1}} \left( t \right)\odot\Theta _{k,j}^{k + 1}$ $\left( {j \in \left[ {1,\left\lceil {\phi b} \right\rceil} \right]} \right)$ denotes the $j$-th aggregated activations' gradients of the client-side $k$-th layer for edge device $i$, ${\bf{z}}_{c,i,{\left\lceil {\phi b} \right\rceil }  + j}^k\left( t \right) = {\bf{z}}_{c,i,{\left\lceil {\phi b} \right\rceil }  + j}^{k + 1}\left( t \right)\odot\Theta _k^{k + 1}$ $\left( {j \in \left[ {1,b - \left\lceil {\phi b} \right\rceil } \right]} \right)$ represents the $j$-th unaggregated activations' gradients of the client-side $k$-th layer for edge device $i$, $\Theta _k^{k + 1}$ is derivative of the $\left(k + 1\right)$-th layer activations with respect to the $k$-th layer activations in the client-side model. $\overline {{\bf{z}}_{c,j}^{{\ell _c}}} \left( t \right)$ and ${{\bf{z}}_{c,i,b}^{\ell _c}\left( t \right)}$ can be calculated by
\begin{align}\label{z_c_j}
\overline {{\bf{z}}_{c,j}^{{\ell _c}}} \left( t \right) = \overline {{\bf{z}}_{s,j}^1} \left( t \right)\odot\Psi,
\end{align}
\begin{align}\label{z_c_i_j}
{\bf{z}}_{c,i,{\left\lceil {\phi b} \right\rceil }  + j}^{{\ell _c}}\left( t \right) = {\bf{z}}_{s,i,{\left\lceil {\phi b} \right\rceil }  + j}^1\odot\Psi,
\end{align}
where $\Psi$ represents derivative of the server-side first layer activations with respect to the client-side ${\ell _c}$-th layer activations. Therefore, for edge device $i$, the client-side model is updated through
\begin{align}\label{stage_7_2_}
{{\bf{W}}_{c,i}}\left( t \right) \leftarrow {{\bf{W}}_{c,i}}\left( t-1 \right) - {\eta _c}{{\bf{G}}_{c,i}}\left( t \right),
\end{align}
where ${\eta}_c$ is the learning rate for the client-side model update.

The EPSL training framework is outlined in~\textbf{Algorithm~\ref{EPSL_procedure}}.

\begin{algorithm}[h]
  \caption{The EPSL training framework}\label{EPSL_procedure}
  \begin{algorithmic}[1]
    \Require
        $b$, ${\eta}_c$, ${\eta}_s$, ${\cal C}$, $\cal{D}$, $\phi$, $f_k$ and $B_k$.
    \Ensure
        ${{\bf{W}}^{\bf{*}}}$
    \State Initialization: ${{\bf{W}}}\left( 0 \right)$
    \For{$t=1,2,...T$ }
    \State
    \State /** {Runs} {on} {edge} {devices} **/
        \ForAll {edge device ${i \in \,{\cal C}}$ in parallel}
        \State ${{\bf{S}}_i}\left( t \right) \leftarrow f\left( {{{\bf{X}}_i}\left( t \right);{{\bf{W}}_{c,i}}\left( t-1 \right)} \right)$
        \State Send $\left( {{\bf{S}}_i\left( t \right),{\bf{y}}_i\left( t \right)} \right)$ to the server
        \EndFor
    \State
    \State /** {Runs} {on} {server} **/
        \State ${\bf{S}}\left( t \right) \leftarrow \left[ {{{\bf{S}}_1}\left( t \right);{{\bf{S}}_2}\left( t \right);...;{{\bf{S}}_C}\left( t \right)} \right] $
        \State ${\bf{y}}\left( t \right) \leftarrow \left[ {{{\bf{y}}_1}\left( t \right);{{\bf{y}}_2}\left( t \right);...;{{\bf{y}}_C}\left( t \right)} \right] $
        \State ${\bf{\hat y}}\left( t \right) \leftarrow f\left( {{\bf{S}}\left( t \right);{{\bf{W}}_s}\left( t-1 \right)} \right) $
        \State Calculate loss function value $L\left( {{\bf{W}}\left( {t - 1} \right)} \right)$
        \State Calculate last-layer aggregated activations' gradients
        \Statex $\;\;\;\;\;\;{\overline {{\bf{z}}_{s,j}^{{\ell _{s}}}} }$
        \State Calculate gradients of server-side model  ${\bf{G}}_s\left( t \right)$
        \State ${{\bf{W}}_s}\left( t \right) \leftarrow {{\bf{W}}_s}\left( t-1 \right) - {\eta}_s {{\bf{G}}_s}\left( t \right)$
        \State Broadcast aggregated activations' gradients in~\eqref{stage_5_1} to
        \Statex $\;\;\;\;\;\;$all edge devices
        \State Send unaggregated activations' gradients in~\eqref{stage_5_1} to c-
        \Statex $\;\;\;\;\;\;$orresponding edge devices
    \State
    \State /** {Runs} {on} {edge} {devices} **/
        \ForAll {edge device ${i \in \,{\cal C}}$ in parallel}
        \State Calculate gradients of client-side model ${{\bf{G}}_{c,i}}\left( t \right)$
        \State ${{\bf{W}}_{c,i}}\left( t \right) \leftarrow {{\bf{W}}_{c,i}}\left( t-1 \right) - {\eta _c}{{\bf{G}}_{c,i}}\left( t \right)$
        \EndFor
    \EndFor
  \end{algorithmic}
\end{algorithm}

\section{Resource Management For EPSL}\label{PROBLEM_FORMULATION}
This section presents the formulation of the resource management problem, with the objective of minimizing the per-round latency. Since the server trains the server-side model after collecting the activations from all participating edge devices, the heterogeneous channel conditions and computing capabilities at the edge devices  result in a severe straggler effect. To decrease the excessive training latency, we devise an effective resource management and layer split strategy that jointly optimizes subchannel allocation, power control, and cut layer selection. For clarity, the decision variables and definitions are listed below.

\begin{itemize}
\item ${\bf{r}}$: $r_i^k \in \left\{ {0,1} \right\}$ denotes the subchannel allocation variable, where $r_i^k=1$ indicates that the $k$-th subchannel is allocated to edge device $i$, and 0 otherwise. The subchannel allocation vector for edge device $i$ is represented as ${{\bf{r }}_i} = \left[ {r _i^1,r _i^2,...,r _i^M} \right]$, where $M$ is the total number of subchannels. Therefore, ${\bf{r}} = \left[ {{{\bf{r}}_1},{{\bf{r}}_2},...,{{\bf{r}}_C}} \right]$ denotes the collection of subchannel allocation decisions.
\item ${\bf{p}}$: $p_k$ represents the transmit power spectral density (PSD) of the $k$-th subchannel. The power control decision is denoted by ${\bf{p}} = \left[ {{p_1},{p_2},...,{p_M}} \right]$.
\item $\bm{\mu}$: $\mu_j \in \left\{ {0,1} \right\}$ is the cut layer selection variable, where $\mu_j=1$ indicates that the $j$-th neural network layer is selected as the cut layer, and 0 otherwise.
$\bm{\mu}  = \left[ {{\mu _1},{\mu _2},...,{\mu _{{\ell_c} + {\ell_s}}}} \right]$ represents the collection of cut layer selection decisions.
\end{itemize}

\subsection{Training Latency Computation}
EPSL is executed in consecutive training rounds until model converges. Without loss of generality, we focus on one training round for analysis. For notational simplicity, index $t$ of training round number is omitted. We conduct the detailed latency analysis of the seven stages in one training round, as follows.

\textit{1) Client-side Model Forward Propagation Latency:}
At this stage, all participating edge devices perform the FP process of client-side models in parallel. Let $\Phi_c^{F}\! \left( \bm{\mu}  \right) = \sum\limits_{j = 1}^{{\ell_c} + {\ell_s}} {{\mu _j}} {\rho _j}$ denote the computation workload (in FLOPs) of the client-side FP process for each data sample, where ${\rho _j}$ is the FP computation workload of propagating the first $j$ layer neural network for one data sample. Since each edge device randomly draw a mini-batch with $b$ data samples to execute the client-side FP process, the edge device $i$'s FP latency is calculated as
\begin{align}\label{client_FP_latency}
T_i^{F} = \frac{{b\,{\kappa _i}\Phi_c^{F}\! \left( \bm{\mu}  \right)}}{{{f_i}}} ,\;\forall i \in \mathcal{C},
\end{align}
where ${{f _i}}$ represents the computing capability (namely, the number of  central processing unit (CPU) cycles per second) of edge device $i$, and ${{\kappa _i}}$ is the computing intensity (i.e., the number of CPU cycles required to complete one float-point operation) of edge device $i$, which is determined by the CPU architecture.

\textit{ 2) Activations Transmission Latency:}
After completing the FP process of the client-side models, each participating edge device sends the activations to the server over the wireless channel. At each training round, we consider frequency-division multiple access for data transmission. Let $\Gamma_s \left( \bm{\mu}  \right) = \sum\limits_{j = 1}^{{\ell_c} + {\ell_s}} {{\mu _j}} {\psi _j}$ represent the data size (in bits) of the activations, where ${\psi _j}$ denotes the data size of the activations at the cut layer $j$. Therefore, the uplink transmission rate from edge device $i$ to the server is given as
\begin{align}\label{smashed_rate}
R_i^U = \sum\limits_{k = 1}^M {r_i^k{B_k}{{\log }_2}\left( {1 + \frac{{{p_k}{G_c}{G_s}\gamma ({F_k},{d_i})}}{{{\sigma ^2}}}} \right)} ,\forall i \in {\cal C},
\end{align}
where ${{B_k}}$ is the bandwidth of the $k$-th subchannel, $G_c$ and $G_s$ denote the effective antenna gain of a edge device and the server, $\gamma(F_k, d_i)$ represents the average channel gain from edge device $i$ to the server, and ${{F_k}}$, ${d_i}$, and ${{\sigma ^2}}$ represent the center frequency of the $k$-th subchannel, the communication distance between edge device $i$ and the server, and the PSD of the noise, respectively. Due to the utilization of the mini-batch with $b$ data samples, the activations transmission latency for edge device $i$ is determined as
\begin{align}\label{smashed_trans_latency}
T_i^{U} = \frac{{b\Gamma_s \left( \bm{\mu}  \right)}}{{R_i^{U}}},\;\forall i \in \mathcal{C}.
\end{align}

\textit{3) Server-side Model Forward Propagation Latency:}
This stage involves executing the server-side FP process\footnote{{The server can execute FP process for multiple edge devices in either a serial or parallel fashion, where the latency in~\eqref{server_FP_latency} would not be affected.}} with activations gathered from all participating edge devices. Let $\Phi _s^{F}\left( \bm{\mu}  \right) = \!\sum\limits_{j = 1}^{{\ell_c} + {\ell_s}} {{{\mu} _j}} \left( {{\rho _{{\ell_c} + {\ell_s}}} - {\rho _j}} \right)$ denote the computation workload of the server-side FP process for one edge device per data sample. Since the server utilizes the activations of all participating edge devices to conduct server-side FP, the computation workload of server is represented as $Cb\Phi _s^{F}\left( \bm{\mu}  \right)$. As a result, the server-side model FP latency can be obtained from
\begin{align}\label{server_FP_latency}
T_s^{F} = \frac{{Cb\,{\kappa _s}\Phi _s^{F}\left( \bm{\mu}  \right)}}{{{f_s}}},
\end{align}
where ${{f _s}}$ denotes the computing capability of the server, and ${{\kappa _s}}$ represents the computing intensity of the server.

\textit{4) Server-side Model Back Propagation Latency\footnote{\rev{ It is noteworthy that the computing workload of the last-layer activations' gradient aggregation is negligible compared to the extensive matrix multiplications during model training, and thus we ignore this part of the latency in our latency analysis.}}:}
After the server-side FP process is completed, EPSL first executes the last-layer BP process to generate the last-layer activations' gradients. Then, EPSL performs last-layer activations' gradient aggregation on the dimension of edge devices. In other words, each edge device employs $\left\lceil {\phi b} \right\rceil$ out of $b$ of its last-layer activations' gradients for aggregation with other edge devices, after which the aggregated activations' gradients go through the BP process. The remaining $\left( {b - \left\lceil {\phi b} \right\rceil } \right)$ unaggregated gradients are directly back-propagated. Let $\Phi _s^{B}\left( \bm{\mu}  \right) = \sum\limits_{j = 1}^{{\ell_c} + {\ell_s}-1} {{\mu _j}} \left( {{\varpi _{{\ell_c} + {\ell_s}-1}} - {\varpi _j}} \right)$ denote the computation workload of the server-side BP process for each data sample (excluding the last layer), where ${\varpi _j}$ is the BP computation workload of propagating the first $j$ layer neural network for one data sample. The computation workload of the last-layer BP process for each data sample is represented as $\Phi _s^L = {\varpi _{{\ell_c} + {\ell_s}}} - {\varpi _{{\ell_c} + {\ell_s} - 1}}$. The computation workload of the last-layer BP is denoted by $Cb\Phi _s^L$, as it is performed on $Cb$ data samples. Similarly, the BP computation workloads for aggregated and unaggregated activations' gradients are represented as $\left\lceil {\phi b} \right\rceil \Phi _s^B\left( \bm{\mu}  \right)$ and $C\left( {b - \left\lceil {\phi b} \right\rceil } \right)\Phi _s^B\left( \bm{\mu}  \right)$, respectively. By ignoring the last-layer aggregation time, the server-side model BP latency is expressed as
\begin{align}\label{server_BP_latency}
T_s^B \!= \!\frac{{\left( {\left\lceil {\phi b} \right\rceil \! + \!C\!\left( {b - \left\lceil {\phi b} \right\rceil } \right)} \right){\kappa _s}\Phi _s^B\left( {\bm{\mu} } \right) + Cb{\kappa _s}\Phi _s^L}}{{{f_s}}}.
\end{align}

\textit{5) Aggregated Activations' Gradients Downlink Broadcasting Latency:}
When the BP process reaches the cut layer, the server broadcasts the aggregated activations' gradients to all participating edge devices over the shared wireless channel. Let $\Gamma_g \left( \bm{\mu}  \right) = \sum\limits_{j = 1}^{{\ell_c} + {\ell_s}} {{\mu _j}} {\chi _j}$ represent the data size of activations' gradients, where the data size of activations' gradients at the cut layer $j$ is denoted by ${\chi _j}$. The downlink broadcasting data rate from the server to every edge device is the same and equal to
\begin{align}\label{server_trans_rate}
{R^B} = \sum\limits_{k = 1}^M {{B_k}{{\log }_2}\left( {1 + \frac{{{p^{DL}}{G_c}{G_s}{\gamma _w}}}{{{\sigma ^2}}}} \right)} ,
\end{align}
where ${\gamma _w} = {\min _{i \in \mathcal{C}, k = 1,2...M}}\left\{ {\gamma ({F_k},{d_i})} \right\}$ represents the weakest average channel gain from edge devices to the server, the transmit PSD of the server is denoted by ${{p^{DL}}}$. Hence, the aggregated activations' gradients downlink broadcasting latency can be obtained by
\begin{align}\label{server_trans_latency}
{T^B} = \frac{{\left\lceil {\phi b} \right\rceil {\Gamma _g}\left( {\bm{\mu} } \right)}}{{{R^B}}}.
\end{align}

\textit{6) Unaggregated Activations' Gradients Transmission Latency:}
After the aggregated activations' gradients downlink broadcasting is completed, the server sends the unaggregated activations' gradients to the corresponding edge devices over the wireless channel. The downlink transmission rate from the server to edge device $i$ is given as
\begin{align}\label{downlink_rate}
R_i^{D} \!=\!\! \sum\limits_{k = 1}^M {r_i^k{B_k}{{\log }_2}\!\!\left( {1 + \frac{{{p^{DL}}G_c G_s { {{\gamma ({F_k},{d_i})}}}}}{{{\sigma ^2}}}} \right)} ,\forall i \in \mathcal{C}.
\end{align}

Therefore, the unaggregated activations' gradients transmission latency for edge device $i$ is determined as
\begin{align}\label{downlink_latency}
T_i^D = \frac{{\left( {b - \left\lceil {\phi b} \right\rceil } \right){\Gamma _g}\left( {\bm{\mu} } \right)}}{{R_i^D}},\;\forall i \in {\cal C}.
\end{align}

\textit{7) Client-side Model Back Propagation Latency:}
At this stage, each edge device performs BP on the client-side model based on the received activations' gradients. Let $\Phi _c^{B}\left( \bm{\mu}  \right) = \sum\limits_{j = 1}^{{\ell_c} + {\ell_s}} {{\mu _j}} {\varpi _j}$ represent the computation workload of the client-side BP process for one data sample. As a result, the client-side BP latency  for edge device $i$  can be obtained from
\begin{align}\label{client_BP_latency}
T_i^{B} = \frac{{b\,{\kappa _i}\Phi _c^{B}\left( \bm{\mu}  \right)}}{{{f_i}}},\;\forall i \in \mathcal{C}.
\end{align}

\subsection{Resource Management Problem Formulation}
As shown in Fig.~\ref{Time_consumption}, the total latency for one training round can be derived as
\begin{align}\label{total_time}
T\!\left( {{\bf{r}},{\bm{\mu }},{\bf{p}}} \right) \!= \!\max_i\! \left\{ {\;\!\!T_i^F\! \!+\! T_i^U\!} \right\}\!+\! T_s^F \!\!+ \!T_s^B \!\!+ \!T^B\! \!\!+\!\max_i \!\left\{ { T_i^D\!\!+\!T_i^B}\right\},
\end{align}

\begin{figure}[t!]
\centering
\includegraphics[width=8.5cm, height=6.2cm]{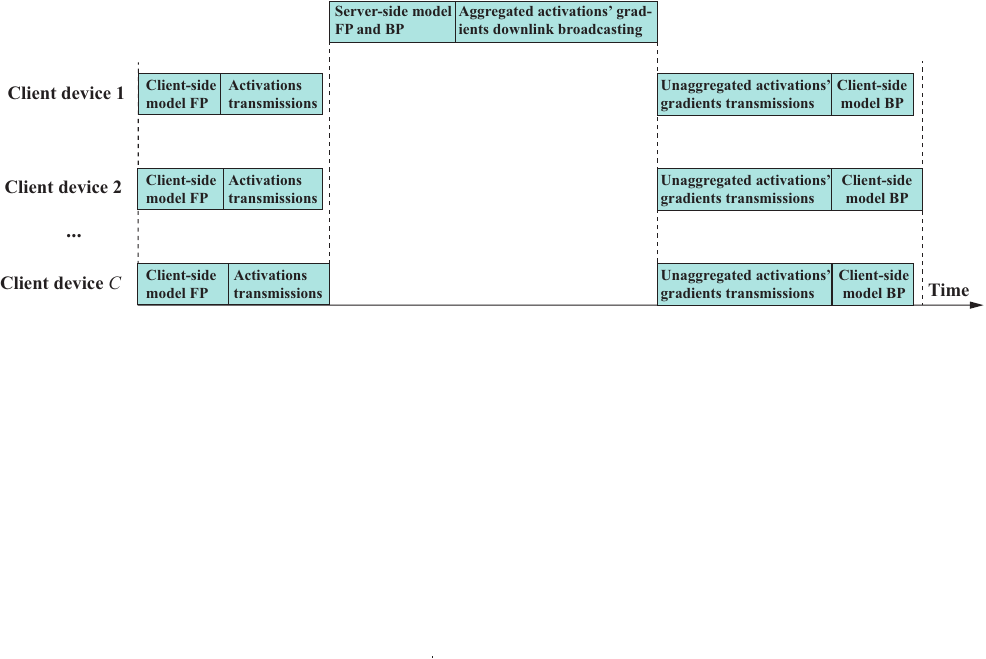}
\vspace{-3.2cm}
\caption{An illustration of EPSL training procedure for one training round.
}
\label{Time_consumption}
\end{figure}

Apparently, the heterogeneous channel conditions and computing capabilities at edge devices may result in  a severe straggler issue. The cut layer selection significantly impacts the communication and computing latency, and appropriate radio resource management is crucial to accelerate the training process. Observing these facts, we formulate the following optimization problem to minimize the per-round latency:
\begin{align}\label{time_minimize_problem}
\mathcal{P}:&\mathop {{\rm{min}}}\limits_{{\bf{r}},{\bm{\mu }},{\bf{p}}} T({\bf{r}},{\bm{\mu }},{\bf{p}})   \\
&\mathrm{s.t.} ~\mathrm{C1:}~r_i^k \in \left\{ {0,1} \right\},\quad \forall i \in {\cal C},k = 1,2,...,M, \nonumber \\
&~\mathrm{C2:}~\sum\limits_{i = 1}^C {r_i^k}  = 1,\quad k = 1,2,...,M, \nonumber \\
&~\mathrm{C3:}~{\mu _j} \in \left\{ {0,1} \right\},\quad j = 1,2,...,{\ell_c} + {\ell_s}, \nonumber\\
&~\mathrm{C4:}~\sum\limits_{j = 1}^{{\ell_c} + {\ell_s}} {{\mu _j}}  = 1, \nonumber \\
&~\mathrm{C5:}~\sum\limits_{k = 1}^M {r_i^k{p_k}{B_k} \le {p_i^{\max }}}, \quad \forall i \in {\cal C}, \nonumber\\
&~\mathrm{C6:}~\sum\limits_{k = 1}^M {\sum\limits_{i = 1}^C {r_i^k{p_k}{B_k}} }  \le {p_{th}}, \nonumber\\
&~\mathrm{C7:}~{p_k} \ge 0,\quad k = 1,2,...,M. \nonumber
\end{align}

Constraints C1 and C2 ensure that each subchannel is solely allocated to one edge device to avoid co-channel interference; C3 and C4 guarantee the uniqueness of cut layer selection, so that the global model is partitioned into the client-side model and the server-side model; C5 is the transmit power constraint of each edge device; C6 illustrates the total uplink transmit power limitation with the threshold $p_{th}$, which is conducive to interference management~\cite{zuo2017power}; C7 ensures that ${p_k}$ is a non-negative value. Note that we consider a stationary network where average link gains vary slowly. Therefore, the cut layer decision, once determined, could last for a long period. We will show the robustness of layer split to channel variation in our simulations. The consideration of mobility for split learning could be left as future work.

Problem~\eqref{time_minimize_problem} is a combinatorial problem with a non-convex mixed-integer nonlinear objective function, which is NP-hard in general. Thus, it is impractical to employ a polynomial-time algorithm to obtain the optimal solution. Therefore, we propose an efficient suboptimal algorithm in the next section.

\section{Solution Approach}\label{Solution_Approach}
To solve problem~\eqref{time_minimize_problem}, we fix the decision variables ${\bm{\mu }}$, ${\bf{p}}$ and consider the subproblem involving subchannel allocation, which is expressed as
\begin{align}\label{sub1}
\mathcal{P}1:&~\mathop {{\rm{min}}}\limits_{{\bf{r}}} T({\bf{r}})   \\
&\mathrm{s.t.} ~\mathrm{C1},~\mathrm{C2},~\mathrm{C5},~\mathrm{C6}. \nonumber
\end{align}

The optimization objective of problem~\eqref{sub1} is still non-convex and contains integer variables,  making it challenging to find the optimal solution. In view of this, we devise a greedy subchannel allocation approach. The core idea of the proposed approach is to allocate subchannel to the straggler with the longest latency of  $T_i^{F}+T_i^{U}$ or $T_i^{D}+T_i^{B}$ (other latency terms are the same for all edge devices). This is because the objective function of problem~\eqref{sub1} is determined by the stragglers' latency in the first and last two stages of one training round, and we prioritize allocating subchannel to the straggler with the longer training latency. To be more specific, we first allocate the subchannel with better propagation characteristics (i.e., lower center frequency) to the edge device with less powerful computing capability. After each edge device attains one channel, the remaining idle channels will be iteratively allocated to the straggler at each iteration until all subchannels are occupied. The procedure of the tailored greedy subchannel allocation approach is described in~\textbf{Algorithm~\ref{greedy-based}}.

\begin{algorithm}[t]
  \caption{Greedy subchannel allocation approach}\label{greedy-based}
  \begin{algorithmic}[1]
    \Require
        $\mathcal{C}$, $F_k$, $B_k$
    \Ensure
        ${\bf{r}^*}$

    \State Initialization: Set  ${\bf{r}} \leftarrow {\bf{0}}$, $\mathcal{A}_1, \mathcal{A}_2 \leftarrow \mathcal{C}$, $\mathcal{K} \leftarrow \left\{ {1,2,...,M} \right\}$
    \For{$j=1,2,...,N$ }
    \State Find $n \leftarrow \arg {\min _{i \in {\cal A}_1}}\left\{ {f_i} \right\}$
    \State Find $m \leftarrow \arg {\min _{k \in \mathcal{K}}}\left\{ {\frac{{{F_k}}}{{{B_k}}}} \right\}$
    \State Let $r_n^m\leftarrow1$, $\mathcal{A}_1 \leftarrow \mathcal{A}_1 - \left\{ n \right\}$, $\mathcal{K}\leftarrow \mathcal{K} - \left\{ m \right\}$
    \State Update $ T_n^{U}$, $ T_n^{D}$
    \EndFor
    \While {$\mathcal{K} \ne \varnothing$}
    \State Find $n_1 \leftarrow \arg {\max _{i \in {\cal A}_2}}\left\{ {T_i^{F} \!+ \!T_i^{U}} \right\}$
    \State Find $n_2 \leftarrow \arg {\max _{i \in {\cal A}_2}}\left\{ {T_i^{D} \!+ \!T_i^{B}} \right\}$
    \State Find $n \leftarrow \arg {\max _{i \in \left\{ {{n_1},{n_2}} \right\}}}\left\{ {T_i^F\! + \!T_i^U\! +\! T_i^D \!+\! T_i^B} \right\}$
    \State Find $m \leftarrow \arg {\min _{k \in \mathcal{K}}}\left\{ {\frac{{{F_k}}}{{{B_k}}}} \right\}$, $r_n^m\leftarrow1$
    \If {${{\bf{r}}_n}$ does not meet C5 in~\eqref{sub1}}
    \State Let $r_n^m\leftarrow0$, $\mathcal{A}_2 \leftarrow \mathcal{A}_2 - \left\{ n \right\}$
    \Else
    \State Update $ T_n^{U}$, $ T_n^{D}$, $\mathcal{K} \leftarrow \mathcal{K} - \left\{ m \right\}$
    \EndIf
    \EndWhile
  \end{algorithmic}
\end{algorithm}

After fixing subchannels according to the aforementioned subchannel allocation approach, we investigate the problem of power control and cut layer selection based on~\eqref{time_minimize_problem}. For notational simplicity, we introduce the auxiliary variable sets ${\bf{\widetilde p}}\! = \left\{ {{p_{1,1}},{p_{1,2}},...,{p_{C,{M_C}}}} \right\}$ and ${\bf{\widetilde B}} \! = \left\{ {{B_{1,1}},{B_{1,2}},...,{B_{C,{M_C}}}} \right\}$ to represent ${\bf{p}}$ and ${\bf{B}}\!  = \left\{ {{B_1},{B_2}...,{B_M}} \right\}$, respectively. The considered problem is then given by

\begin{align}\label{sub2}
&~\mathop {{\rm{min}}}\limits_{{\bm{\mu }},{{{\bf{\widetilde p}}}}} T({\bm{\mu }},{{\bf{\widetilde p}}})    \\
&\mathrm{s.t.} ~\mathrm{C3},~\mathrm{C4} \nonumber \\
&~\mathrm{\widetilde{C}5:}~\sum\limits_{\xi  = 1}^{{M_i}} {{p_{i,\xi }}{B_{i,\xi }} \le p_i^{\max }}, \quad \forall i \in {\cal C}, \nonumber\\
&~\mathrm{\widetilde{C}6:}~\sum\limits_{i = 1}^C {\sum\limits_{\xi  = 1}^{{M_i}} {{p_{i,\xi }}{B_{i,\xi }}} }  \le {p_{th}}, \nonumber\\
&~\mathrm{\widetilde{C}7:}~{p_{i,\xi }} \ge 0,\quad \forall i \in {\cal C}, \xi=1,2,...,M_i, \nonumber
\end{align}
where $M_i$ represents the number of subchannels allocated to edge device $i$, ${p_{i,\xi }}$ denotes the transmit PSD on the $\xi$-th subchannel for edge device $i$, and ${B_{i,\xi }}$ is the bandwidth of $\xi$-th subchannel for edge device $i$.

Problem~\eqref{sub2} is a mixed-integer nonlinear programming. The challenges in addressing this problem stem primarily from the nonconvexity of the objective function. In order to linearize the objective function, we introduce the auxiliary variables ${T_1}$ and ${T_2}$ that satisfy ${T_1} \ge \mathop {\max }\limits_i \left\{ {\;T_i^F + T_i^U} \right\}$ and ${T_2} \ge \mathop {\max }\limits_i \left\{ {T_i^D + T_i^B} \right\}$, respectively. The problem~\eqref{sub2} can be converted into
\begin{align}\label{sub2_convert}
&~\mathop {{\rm{min}}}\limits_{{\bm{\mu }},{\bf{\widetilde p}},{T_1},{T_2}} \widetilde{T}({\bm{\mu }},{\bf{\widetilde p}},{T_1},{T_2})\! \\
&\mathrm{s.t.} ~\mathrm{C3},~\mathrm{C4},~\mathrm{\widetilde{C}5},~\mathrm{\widetilde{C}6},~\mathrm{\widetilde{C}7} \nonumber \\
&~\mathrm{C8:}~\frac{{b{\mkern 1mu} {\kappa _i}\!\!\!\sum\limits_{j = 1}^{{\ell_c} + {\ell_s}} {\!\!\!{\mu _j}} {\rho _j}}}{{{f_i}}}\! +\!\frac{{b\!\!\sum\limits_{j = 1}^{{\ell_c} + {\ell_s}} {\!\!{\mu _j}} \psi_j }}{{\sum\limits_{\xi  = 1}^{{M_i}} \!{\!{B_{i,\xi }}} {\rm{lo}}{{\rm{g}}_2}\!\left( {1\! + \!\frac{{{p_{i,\xi }}G_c G_s { {\gamma ({F_k},{d_i})} }}}{{{\sigma ^2}}}} \right)}} \!\le\! {T_1},\nonumber\\
&{\kern 225pt}\forall i \in {\cal C}, \nonumber\\
&~\mathrm{C9:}~\frac{{\left( {b - \left\lceil {\phi b} \right\rceil } \right)\sum\limits_{j = 1}^{{\ell _c} + {\ell _s}} {{\mu _j}} {\chi _j}}}{{R_i^D}} + \frac{{b{\kappa _i}\sum\limits_{j = 1}^{{\ell _c} + {\ell _s}} {{\mu _j}} {\varpi _j}}}{{{f_i}}} \le {T_2},\quad \!\!\forall i \in {\cal C}, \nonumber
\end{align}
where $\widetilde{T}({\bm{\mu }},{\bf{\widetilde p}},{T_1},{T_2})\!={{T_1} + T_s^F + T_s^B + T^B + {T_2}}$, the additional constraint C8 and C9 are established based on ${T_1} \ge \mathop {\max }\limits_i \left\{ {\;T_i^F + T_i^U} \right\}$ and ${T_2} \ge \mathop {\max }\limits_i \left\{ {T_i^D + T_i^B} \right\}$.

The transformed problem~\eqref{sub2_convert} is equivalent to the original problem~\eqref{sub2}, as the optimal solution ${T_1^*}$ and ${T_2^*}$ obtained from problem~\eqref{sub2_convert} must meet $T_1^* = \mathop {\max }\limits_i \left\{ {\;T_i^F + T_i^U} \right\}$ and $T_2^* = \mathop {\max }\limits_i \left\{ {\;T_i^D + T_i^B} \right\}$. Otherwise, there must be ${T'_1}$ and ${T'_2}$ satisfying $\mathop {\max }\limits_i \left\{ {\;T_i^F + T_i^U} \right\} \le {{T'}_1} < T_1^*$ and $\mathop {\max }\limits_i \left\{ {\;T_i^D + T_i^B} \right\} \le {{T'}_2} < T_2^*$ with lower objective function value. Even though the objective function of problem~\eqref{sub2_convert} is linear, the nonconvexity of constraint C8 makes solving the problem challenging. Therefore, we introduce a set of auxiliary variables $\bm{\theta}   = \left\{ {{\theta _{1,1}},{\theta _{1,2}},...,{\theta _{N,{M_N}}}} \right\}$, which are determined as
\begin{align}\label{substitution variables}
{\theta _{i,\xi }} \!=\! {B_{i,\xi }}{\rm{lo}}{{\rm{g}}_2}\!\!\left( {\!1 \!+ \!\frac{{{p_{i,\xi }}G_c G_s {\gamma ({F_k},{d_i})}}}{{{\sigma ^2}}}}\right)\!.
\end{align}

We derive the feasible region of auxiliary variables as ${\theta_{i,\xi }} \ge 0$  based on transmit PSD's feasible region ${p_{i,\xi }} \ge 0$. By substituting the variable $\bm{\theta}$ and its corresponding feasible region into the problem~\eqref{sub2_convert}, the further transformed problem is given as
\begin{align}\label{sub2_further_convert}
&~\mathop {{\rm{min}}}\limits_{{\bm{\mu }},{\bm{\theta}},{T_1},{T_2}} \widetilde{T}({\bm{\mu }},{\bm{\theta}},{T_1},{T_2})  \\
&\mathrm{s.t.} ~\mathrm{C3},~\mathrm{C4},~\mathrm{C9} \nonumber \\
&~\mathrm{\hat{C}5:}~\sum\limits_{\xi  = 1}^{{M_i}} {{\sigma ^2}{B_{i,\xi }}\frac{{{2^{\frac{{{\theta _{i,\xi }}}}{{{B_{i,\xi }}}}}} - 1}}{{G_c G_s {\gamma ({F_k},{d_i})}}} \le p_i^{\max }},  \quad \forall i \in {\cal C}, \nonumber\\
&~\mathrm{\hat{C}6:}~\sum\limits_{i = 1}^C {\sum\limits_{\xi  = 1}^{{M_i}} {{\sigma ^2}{B_{i,\xi }}\frac{{{2^{\frac{{{\theta _{i,\xi }}}}{{{B_{i,\xi }}}}}} - 1}}{{G_c G_s {\gamma ({F_k},{d_i})}}}} }  \le {p_{th}}, \nonumber\\
&~\mathrm{\hat{C}7:}~{\theta_{i,\xi }} \ge 0,\quad \forall i \in {\cal C},\xi=1,2,...,M_i, \nonumber\\
&~\mathrm{\hat{C}8:}~\!\frac{{b{\kappa _i}\!\!\sum\limits_{j = 1}^{{\ell_c} + {\ell_s}} \!\!\!{{\mu _j}} {\rho _j}}}{{{f_i}}}\! + \!\frac{{b\!\!\sum\limits_{j = 1}^{{\ell_c} + {\ell_s}} \!\!\!{{\mu _j}} \psi_j }}{{\sum\limits_{\xi  = 1}^{{M_i}} {{\theta _{i,\xi }}} }} \!\!\le {T_1},\quad\forall i \in {\cal C}, \nonumber
\end{align}

Although problem~\eqref{sub2_further_convert} is still non-convex, we observe that if we fix the decision variables $\bm{\mu}$, ${T_1}$ and ${T_2}$, it can be written as
\begin{align}\label{sub2_sub1}
\mathcal{P}2:&~\mathop {{\rm{min}}}\limits_{{\bm{\theta}}} \widetilde{T}({\bm{\theta}})    \\
&\mathrm{s.t.} ~\mathrm{\hat{C}5},~\mathrm{\hat{C}6},~\mathrm{\hat{C}7},~\mathrm{\hat{C}8} \nonumber
\end{align}

Problem~\eqref{sub2_sub1} is obviously a convex problem with respect to ${\bm{\theta}}$, which can be efficiently solved through available toolkits such as CVX~\cite{CVX}.

Similarly, if the decision variables ${\bm{\theta}}$, ${T_1}$ and ${T_2}$ are fixed, problem~\eqref{sub2_further_convert} is then converted to
\begin{align}\label{sub2_sub2}
\mathcal{P}3:&~\mathop {{\rm{min}}}\limits_{{\bm{\mu }}} \widetilde{T}({\bm{\mu }})    \\
&\mathrm{s.t.} ~\mathrm{C3},~\mathrm{C4},~\mathrm{\hat{C}8},~\mathrm{C9}, \nonumber
\end{align}

Fortunately, problem~\eqref{sub2_sub2} is a standard mixed integer linear programming (MILP). We employ the exhaustive search algorithm to address the problem~\eqref{sub2_sub2}. It is noted that the dimension of $\mu_j$ is the number of layers of a neural network, which is typically a small number. For example, the well-known image classification neural network AlexNet~\cite{krizhevsky2017imagenet} and GoogLeNet~\cite{szegedy2015going} contain 8 and 22 neural network layers. Therefore, exhaustive search can solve the problem efficiently.


By fixing the decision variables $\bm{\mu}$ and ${\bm{\theta}}$, we convert problem~\eqref{sub2_further_convert} into a linear programming with respect to ${T_1}$ and ${T_2}$, which is given by
\begin{align}\label{sub2_sub3}
\mathcal{P}4:&~\mathop {{\rm{min}}}\limits_{{{T_1},{T_2}}} \widetilde{T}({T_1},{T_2})   \\
&\mathrm{s.t.} ~\mathrm{\hat{C}8},~\mathrm{{C}9} \nonumber
\end{align}

We can easily obtain the optimal solution by observing problem~\eqref{sub2_sub3}, i.e.
\begin{align}\label{sub2_sub3_optimal_1}
&T_1^* = \mathop {\max }\limits_i \Bigg\{ \frac{{b{\kappa _i}{\rho _j}}}{{{f_i}}} + \frac{{b{\psi _j}}}{{\sum\limits_{\xi  = 1}^{{M_i}} {{\theta _{i,\xi }}} }}\Bigg\} \;,\\
&T_2^* = \mathop {\max }\limits_i\left\{ {\frac{{\left( {b - \left\lceil {\phi b} \right\rceil } \right){\chi _j}}}{{R_i^D}} + \frac{{b{\kappa _i}{\varpi _j}}}{{{f_i}}}} \right\},
\end{align}

As aforementioned, we decompose the original problem~\eqref{time_minimize_problem} into four less complicated subproblems (i.e., $\mathcal{P}1$, $\mathcal{P}2$, $\mathcal{P}3$, $\mathcal{P}4$) based on different decision variables, and present efficient algorithms to address each subproblem. Therefore, we propose a block-coordinate descent (BCD)-based algorithm~\cite{tseng2001convergence} to tackle problem~\eqref{time_minimize_problem}, which is described in~\textbf{Algorithm~\ref{BCD_based}}, where $\bm{r}^{\tau}$, $\bm{\mu}^{\tau}$, $\bm{\theta}^{\tau}$, $T_{1}^{\tau}$, and $T_{2}^{\tau}$ denote $\bm{r}$, $\bm{\mu}$, $\bm{\theta}$, $T_{1}$, and $T_{2}$ at the $\tau$-iteration. \rev{Despite of the lack of theoretically guaranteed convergence due to the non-convex and mixed-integer nature~\cite{grippo2000convergence} of the original problem $\mathcal{P}$, the proposed BCD-based  algorithm exhibits notably efficient and reliable performance in practice, i.e., it converges within only a few iterations and  to the same solution under various initial conditions~\cite{chen2021qos,chen2019resonant,gu2018coordinate}.}

\begin{algorithm}[h]
  \caption{BCD-based Algorithm.}\label{BCD_based}
  \begin{algorithmic}[1]
    \Require
        convergence tolerance $\varepsilon $, iteration index $\tau  = 0$.
    \Ensure
        ${\bf{r}}^*$, ${\bm{\mu }^*}$, ${\bf{p}^*}$
    \State Initialization: ${\bf{r}}^0$, ${\bm{\mu }^0}$, ${{\bm{\theta }}^0}$, $T_1^0$, $T_2^0$
    \Repeat
    \State $\tau \leftarrow \tau+1$
    \State Update ${\bf{r}}^{\tau}$ based on~\textbf{Algorithm~\ref{greedy-based}}
    \State Update ${\bm{\theta }^{\tau}}$ by solving convex problem~\eqref{sub2_sub1}
    \State Update ${\bm{\mu }^{\tau}}$ by solving MILP~\eqref{sub2_sub2}
    \State Update ${T_1^{\tau}}$ and ${T_2^{\tau}}$ according to equation~\eqref{sub2_sub3_optimal_1} and (34)
    \Until $| {\widetilde{T}({\bf{r}}^{\tau}\! ,{\bm{\mu }^{\tau}}\!,{\bm{\theta }^{\tau}}\!,{T_1^{\tau}},{T_2^{\tau}}) - \widetilde{T}({\bf{r}}^{\tau-1}\!\!, {\bm{\mu }^{\tau-1}}\!\!, {\bm{\theta }^{\tau-1}}\!, {T_1^{\tau-1}}\!,{} } $ ${T_2^{\tau-1}} ) | \le \varepsilon $

  \end{algorithmic}
\end{algorithm}

\section{Simulation Results}\label{simulation results}
This section provides numerical results to evaluate the learning performance of the proposed EPSL framework and the effectiveness of the tailored resource management and layer split strategy.
\subsection{Simulation Setup}
In the simulations, we deploy $C$ edge devices uniformly distributed within the coverage radius $d_{max}=200$m, with server in its center. $C$ is set to 5 by default unless specified otherwise. The computing capability of each edge device is uniformly distributed within $[1, 1.6]\times {10^9}$ cycles/s, and the computing capability of the server is set to $5 \times {10^9}$ cycles/s. The total available bandwidth is 200 MHz, with equal subchannel bandwidth ${B_1} = ... = {B_M} = B =$ 10MHz. The transmit PSD of the server $p^{DL}=-50$dBm/Hz, and the PSD of the noise is set to $-174$dBm/Hz~\cite{hu2019edge,lin2021spatial,lin2022tracking}. we consider the maximum transmit power $p_1^{\max } = ... = p_C^{\max } = {p^{\max }}=$31.76dBm, and an equal computing intensity ${\kappa _1} = ... = {\kappa _C} = {\kappa } =\frac{1}{{16}}$ cycles/FLOPs. We adopt the channel model in~\cite{samimi2015probabilistic}, where the average path loss exponents for LoS and NLoS are 2.1 and 3.4, and the standard deviations of shadow fading for LoS and NLoS are 3.6dB and 9.7dB. For readers' convenience, the detailed simulation parameters are summarized in Table 3.

\begin{table}[t]\label{table_2}
  \centering
  \caption{Simulation Parameters.}
  \renewcommand{\arraystretch}{1.25}{
  \setlength{\tabcolsep}{0.5mm}{
\begin{tabular}{|c|c|c|c|}
\hline
\textbf{Parameter}          & \textbf{value} & \textbf{Parameter} & \textbf{value}  \\ \hline
$f_s$             & $5 \times {10^9} $cycles/s              & $f_i$                 & $[1, 1.6]\times {10^9} $cycles/s                        \\ \hline
$C$             & 5              & $M$           & 20                    \\ \hline
$p^{DL}$               & $-50$dBm/Hz              & $G_cG_s$                  & 10                        \\ \hline
$\sigma^2$        & $-174$dBm/Hz            & $b$             & 64                       \\ \hline
$\kappa _s$        & $\frac{1}{{32}}$ cycles/FLOPs            & $\kappa$             & $\frac{1}{{16}}$ cycles/FLOPs                       \\ \hline
$d_{max}$        & $200$m            &$B$            &$10$MHz                        \\ \hline
$\eta _c$        & $1.5 \times {10^{-4}} $            &$\eta _s$            &$ {10^{-4}} $                        \\ \hline
$p^{\max }$        & $31.76$dBm            &$p_{th}$            &$36.99$dBm                         \\ \hline
\end{tabular}}}
\end{table}

\begin{figure}[t!]
\centering
\includegraphics[width=6.5cm]{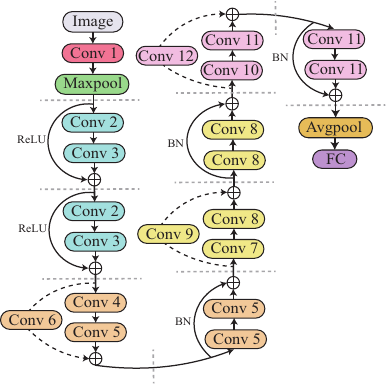}
\vspace{0cm}
\caption{ResNet-18 network structure, where the gray dashed line marks the potential choice of the cut layer.
}
\label{NN_formal}
\end{figure}

\begin{table}[h]
  \centering
  \caption{ResNet-18 Network Parameters.}
  \renewcommand{\arraystretch}{1.2}{
  \setlength{\tabcolsep}{0.5mm}{
\begin{tabular}{|c|c|c|c|c|c|}
\hline
\multirow{2}{*}{\textbf{Layer name}} &
  \multirow{2}{*}{\textbf{NN unit}} &
  \multirow{2}{*}{\textbf{Activation}} &
  \multirow{2}{*}{\textbf{\begin{tabular}[c]{@{}c@{}}Layer size\\ (MB)\end{tabular}}} &
  \multirow{2}{*}{\textbf{\begin{tabular}[c]{@{}c@{}}FP FLOPs\\ (MFLOP)\end{tabular}}} &
  \multirow{2}{*}{\textbf{\begin{tabular}[c]{@{}c@{}}Activations\\ (MB)\end{tabular}}} \\
        &         &         &        &        &          \\ \hline
CONV1   & 7$\times$7,64  & BN+ReLU & 0.0364 & 9.8304 & 0.25     \\ \hline
CONV2   & 3$\times$3,64  & BN+ReLU & 0.1411 & 9.5027 & 0.0625   \\ \hline
CONV3   & 3$\times$3,64  & BN      & 0.1414 & 9.4863 & 0.0625   \\ \hline
CONV4   & 3$\times$3,128 & BN      & 0.2827 & 4.7432 & 0.0313   \\ \hline
CONV5   & 3$\times$3,128 & BN      & 0.564  & 9.4618 & 0.0313   \\ \hline
CONV6   & 1$\times$1,128 & BN      & 0.0327 & 0.5489 & 0.0313   \\ \hline
CONV7   & 3$\times$3,256 & BN      & 1.1279 & 4.7309 & 0.0156   \\ \hline
CONV8   & 3$\times$3,256 & BN      & 2.2529 & 9.4495 & 0.0156   \\ \hline
CONV9   & 1$\times$1,256 & BN      & 0.1279 & 0.5366 & 0.0156   \\ \hline
CONV10  & 3$\times$3,512 & BN      & 4.5059 & 4.7247 & 0.0078   \\ \hline
CONV11  & 3$\times$3,512 & BN      & 9.0059 & 9.4433 & 0.0078   \\ \hline
CONV12  & 1$\times$1,512 & BN      & 0.5059 & 0.5304 & 0.0078   \\ \hline
MAXPOOL & 3$\times$3,64  & /       & /      & 0.0655 & 0.0625   \\ \hline
AVGPOOL & 2$\times$2     & /       & /      & /      & 0.0020   \\ \hline
FC      & 7       & /       & 0.0137 & 0.0036 & 2.67E-05 \\ \hline
    \end{tabular}}}%
  \label{tab:addlabel}%
\end{table}%

We adopt the image classification datasets MNIST~\cite{lecun1998mnist} and HAM10000~\cite{tschandl2018ham10000} to evaluate EPSL's learning performance. The MNIST dataset contains grayscale images with handwritten digits 0 to 9, whereas HAM10000 dataset includes seven different types of skin images, such as melanoma and dermatofibroma. The MNIST dataset has 60000 training samples and 10000 test samples, and HAM10000 dataset has 8000 training samples and 2015 test samples. Furthermore, we conduct experiments under IID and non-IID data settings. The data samples are shuffled and distributed evenly to all edge devices in the IID setting. In the non-IID setting~\cite{chen2022federated,yang2021achieving}, each edge devices has training samples only associated with two categories.

\begin{table}[t]\label{table11}
\caption{The Converged Test Accuracy for HAM10000 under IID Setting.}
  \renewcommand{\arraystretch}{1.2}{
  \setlength{\tabcolsep}{2.5mm}{
\begin{tabular}{|l|c|c|c|c|}
\hline
\diagbox{\textbf{\begin{tabular}[c]{@{}c@{}}Learning\\ framework  \end{tabular}}}{\textbf{$\bm{C}$}} & \textbf{5} & \textbf{10} & \textbf{15} \\ \hline
\textbf{\quad \quad \quad vanilla SL} & 77.58$\%$  & 77.65$\%$  & 77.62$\%$   \\ \hline
\textbf{\quad \quad \quad SFL} & 77.37$\%$  & 77.21$\%$  & 77.24$\%$   \\ \hline
\textbf{\quad \quad \quad PSL$($EPSL with $\phi=0)$} & 77.33$\%$  & 77.29$\%$  & 77.12$\%$   \\ \hline
\textbf{\quad \quad \quad EPSL$(\phi=0.5)$} & 77.24$\%$  & 77.11$\%$  & 76.93$\%$  \\ \hline
\textbf{\quad \quad \quad EPSL$(\phi=1)$} & 77.12$\%$  & 76.37$\%$ & 74.29$\%$  \\ \hline
\end{tabular}}}
\end{table}

We deploy the ResNet-18 network~\cite{he2016deep}, which primarily consists of convolution (CONV) layers, pooling (POOL) layers, and  fully-connected (FC) layers. The network structure and network parameters of the implemented  ResNet-18  are detailed in Fig.~\ref{NN_formal} and Table 4. We resize the input images to 64$\times$64 to fit the neural network's input. The mini-batch size is set to 64.

\subsection{Performance Evaluation of the Proposed EPSL Framework}

In this subsection, we conduct the performance evaluation of the proposed EPSL framework in terms of test accuracy, convergence speed and training latency. To investigate the advantages of the EPSL framework, we compare it with three well-known SL frameworks:

\begin{itemize}
\item \textbf{Vanilla SL:} Vanilla SL sequentially trains the client-side model across all participating edge devices. Specifically, model training is solely conducted on a single edge device and server. After finishing the training for one edge device, the updated client-side model is transferred to the next edge device for model training~\cite{vepakomma2018split}.
\item \textbf{SFL:} SFL trains the client-side model on all participating edge devices in a parallel fashion, and then transmit the updated client-side model to the server for model aggregation~\cite{thapa2022splitfed}.
\item \textbf{PSL:} PSL trains the shared server-side model using all edge devices data, while training client-side models based on their respective data. PSL is the special case of EPSL with $\phi=0$~\cite{kim2022bargaining,joshi2021splitfed}.
\item \textbf{EPSL with phased training (EPSL-PT):} The phased training is a hybrid distributed training framework that first leverages EPSL with $\phi=1$ initially, and then switches to employing PSL (i.e., EPSL with $\phi=0$) for model training. This demostrates the flexibility of the proposed EPSL via appropriate control of $\phi$.
\end{itemize}

\begin{figure}[t]
\centering
\begin{subfigure}{0.5\textwidth}
  \centering
  \includegraphics[width=6.7cm,height=5.7cm]{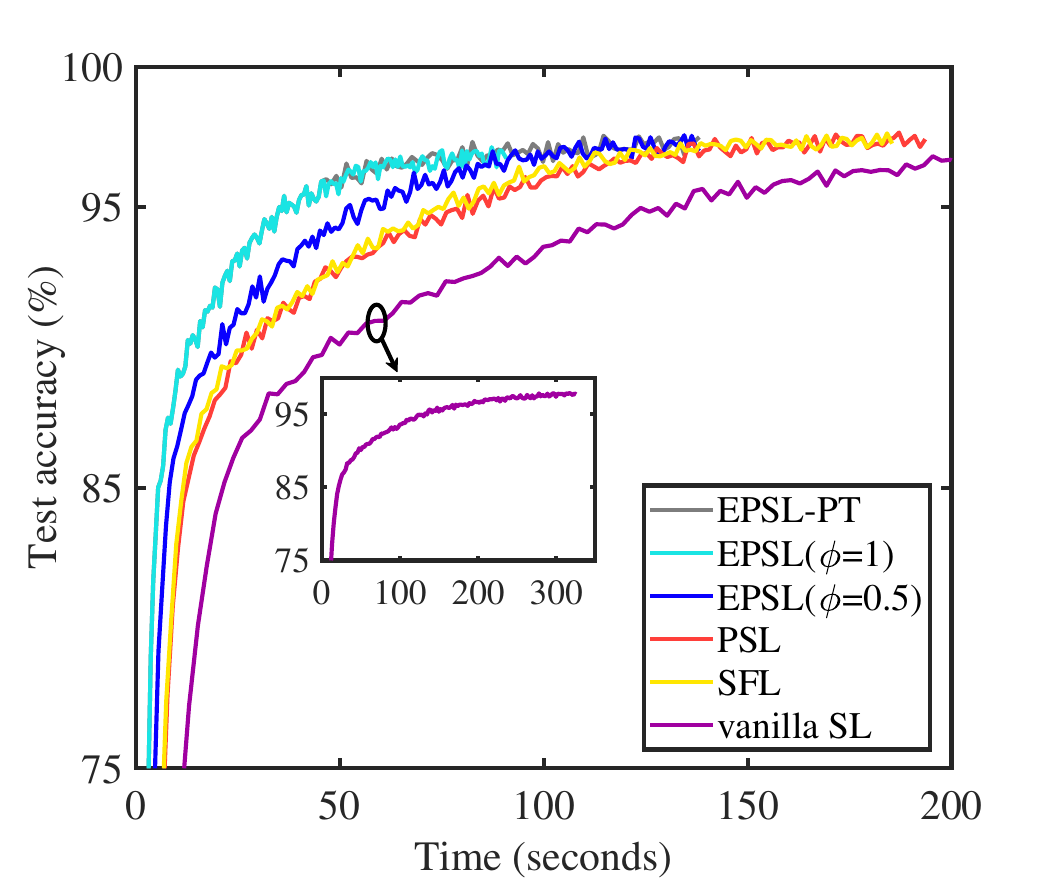}
  \caption{Test accuracy on MNIST under IID setting.}
  \label{test_accuracy_1a}
\end{subfigure}\\
\begin{subfigure}{0.5\textwidth}
  \centering
  \includegraphics[width=6.7cm,height=5.7cm]{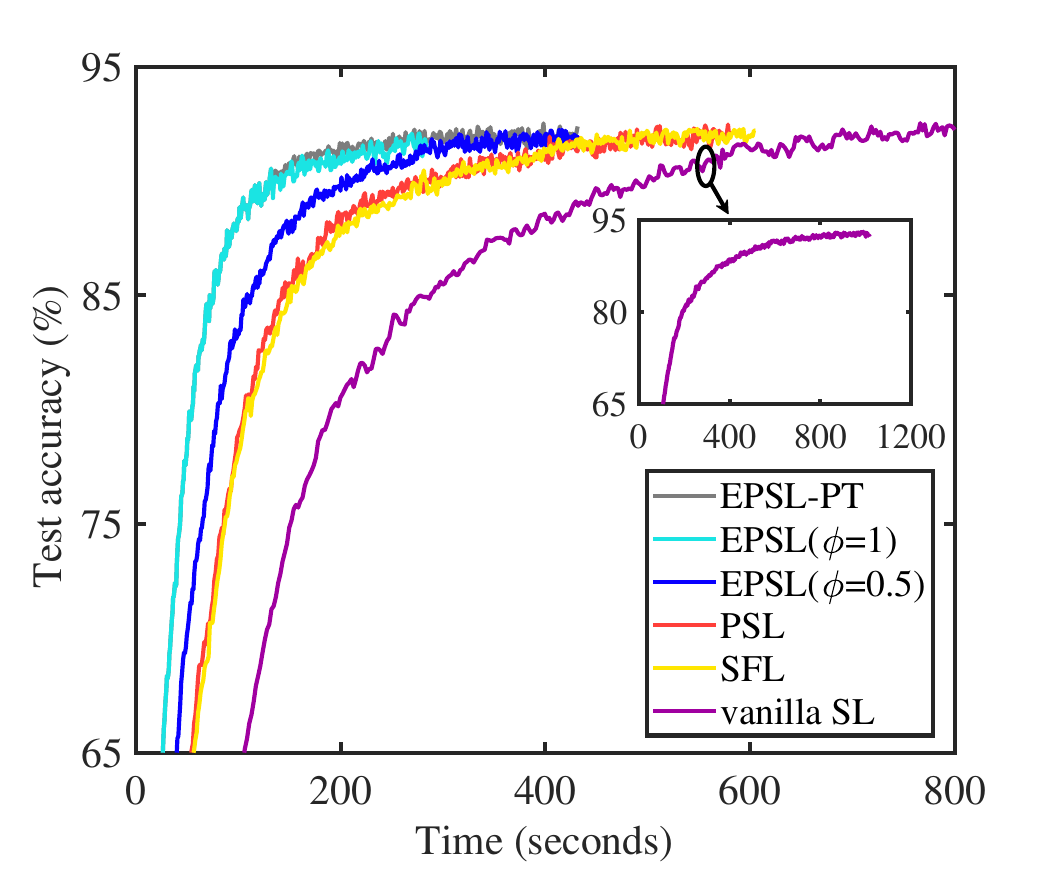}
  \caption{Test accuracy on MNIST under IID setting.}
  \label{test_accuracy_1b}
\end{subfigure}
\caption{The test accuracy of vanilla SL, SFL, PSL and EPSL on MNIST under IID and non-IID settings using ResNet-18.}
\label{test_accuracy_1}
\end{figure}

Table 5 shows the converged test accuracy for HAM10000 dataset under IID setting, where the accuracy is taken at the 300-th epoch (one epoch means going through all local data) when these learning frameworks all converges. It can be seen that EPSL with $\phi=0.5$ and $\phi=1$ achieve a similar learning accuracy as compared with vanilla SL, SFL and PSL when the number of edge devices is small. When the number of edge devices is large, EPSL with $\phi=1$ cannot converge well, because the aggregated activations' gradients could be too "general" to fit each local dataset. However, as we decrease aggregation ratio $\phi$, the performance again becomes comparable to vanilla SL, SFL and PSL. This demonstrates the effectiveness of last-layer activations' gradient aggregation and the flexibility of our scheme via appropriate control of $\phi$ (like EPSL-PT that switches to $\phi=0$ when the model is close to the convergence region).

\begin{figure}[t]
\centering
\begin{subfigure}{0.5\textwidth}
  \centering
  \includegraphics[width=6.7cm,height=5.7cm]{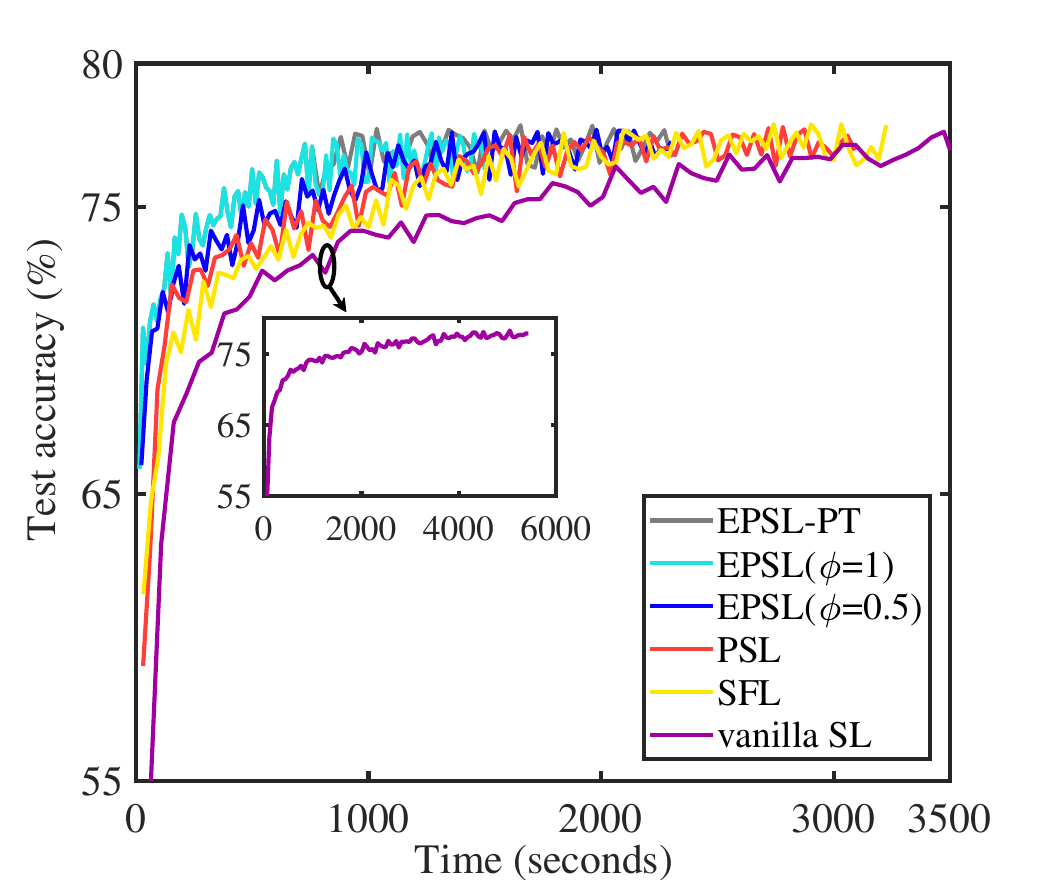}
  \caption{Test accuracy on HAM10000 under IID setting.}
  \label{test_accuracy_a}
\end{subfigure}\\
\begin{subfigure}{0.5\textwidth}
  \centering
  \includegraphics[width=6.7cm,height=5.7cm]{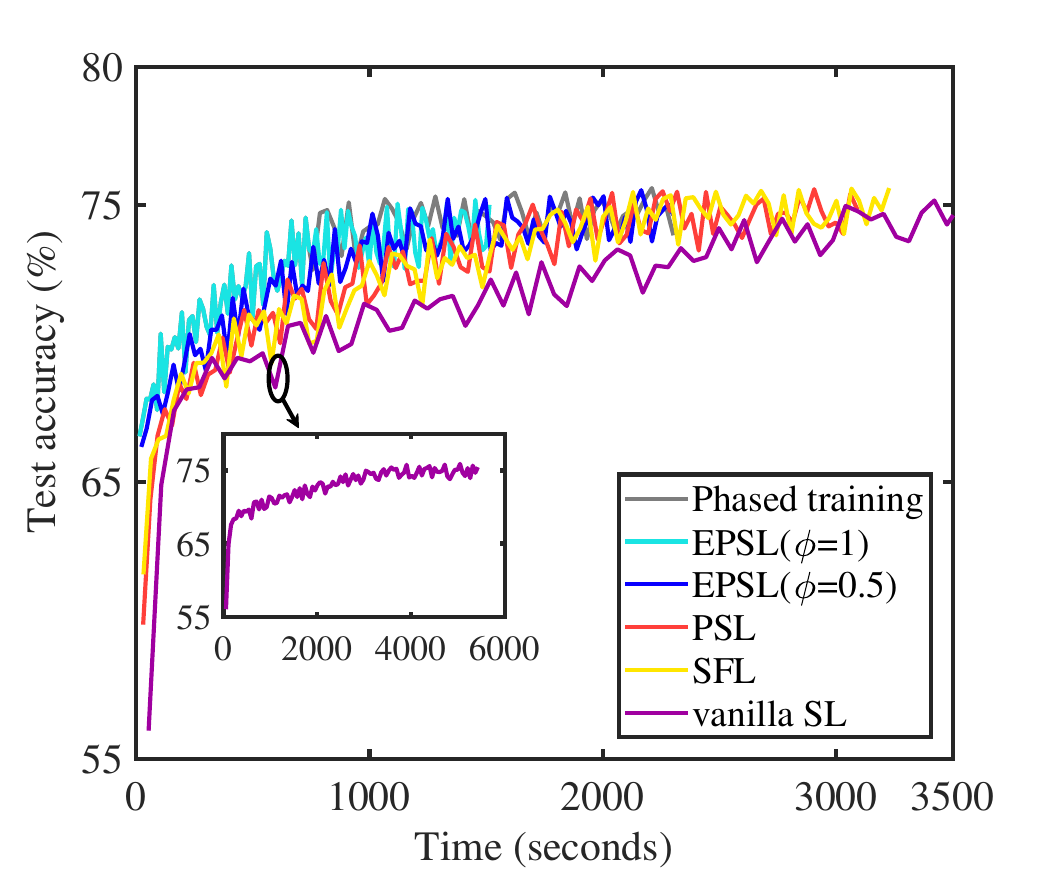}
  \caption{Test accuracy on HAM10000 under IID setting.}
  \label{test_accuracy_b}
\end{subfigure}
\caption{The test accuracy of vanilla SL, SFL, PSL and EPSL on HAM10000 under IID and non-IID settings using ResNet-18.}
\label{test_accuracy}
\end{figure}

Fig.\ref{test_accuracy_1}-\ref{test_accuracy} present the test accuracy of different SL frameworks on MNIST and HAM10000 datasets using ResNet-18. It is seen that EPSL retains similar test accuracy as vanilla SL, SFL and PSL as the models converge. Furthermore, EPSL demands the lowest time budget to achieve a target accuracy. The reason is twofold: one is that EPSL conducts the last-layer gradient aggregation on the server-side model to drastically reduce the dimension of activations' gradient and the server's computation workload in the BP process,  and the other is that EPSL eliminates the necessity for model exchange. Fig.~\ref{test_accuracy_1b} and Fig.~\ref{test_accuracy_b} show that the convergence speed of the four SL frameworks is slower under non-IID setting than under IID setting.

Fig.~\ref{client_num} illustrates the total training latency for achieving target accuracy (75.5$\%$) versus the number of edge devices under different SL frameworks. The total training latency of vanilla SL increases as the number of edge devices grows. This is due to the fact that, with the fixed total available bandwidth, the bandwidth allocated to each edge device reduces considerably, resulting in lower uplink transmission rate. Different from vanilla SL, the training latency of SFL, PSL and EPSL gradually decreases with the growth of the number of edge devices. The reason is that each edge device is assigned with fewer number of data samples while the number of edge devices increase. Moreover, SFL, PSL and EPSL employ the parallel client-side model training schemes to accelerate model training, which leads to lower training latency. The EPSL is the SL framework with the lowest training latency, and its training latency decreases faster as the number of edge devices increases.  This is because that the last-layer gradient aggregation reduces the data size of the transmitted data, thereby mitigating the adverse impact of the wireless spectrum scarcity on training latency. It demonstrates that EPSL is more capable than vanilla SL, SFL, and PSL of enabling low-latency model training over unfavourable wireless communication conditions.

\begin{figure}[t!]
\centering
\includegraphics[width=6.7cm,height=5.7cm]{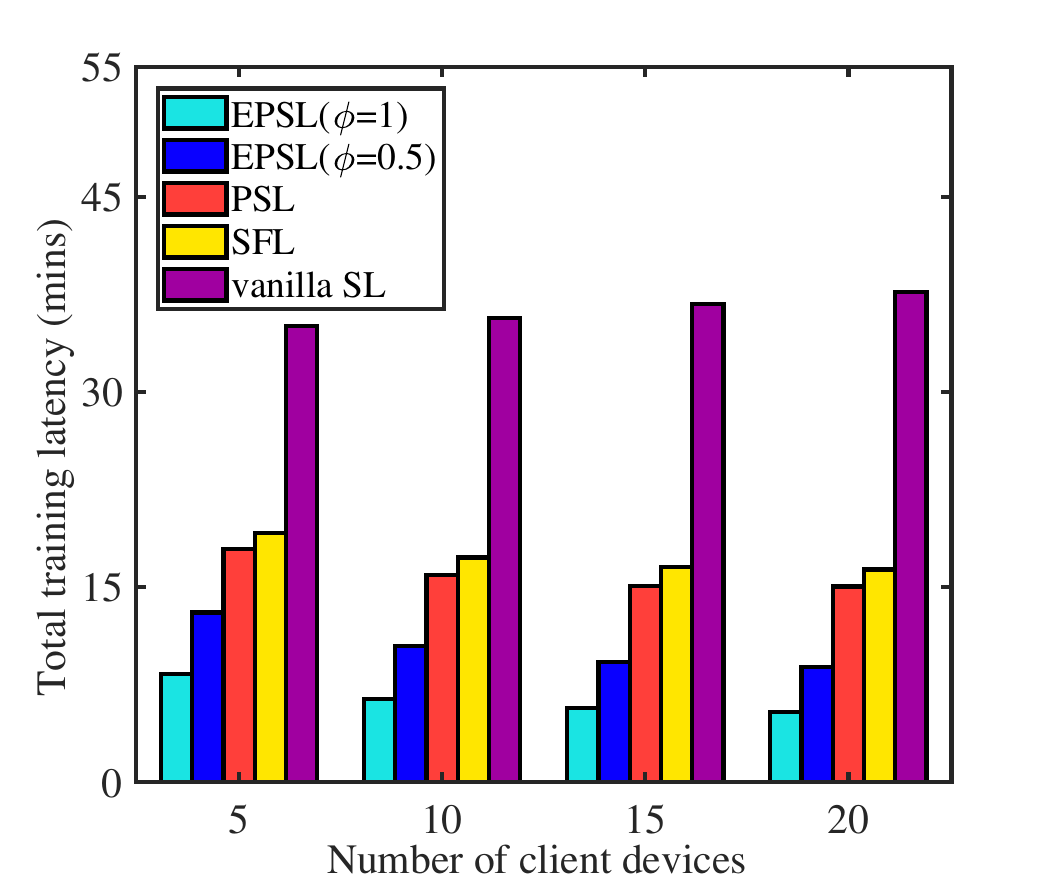}
\vspace{0cm}
\caption{The total training latency for achieving target accuracy on HAM10000 versus the number of edge devices under IID setting with $M$=20 and $D$=8000.
}
\label{client_num}
\end{figure}

\begin{figure}[t!]
\centering
\includegraphics[width=6.7cm,height=5.8cm]{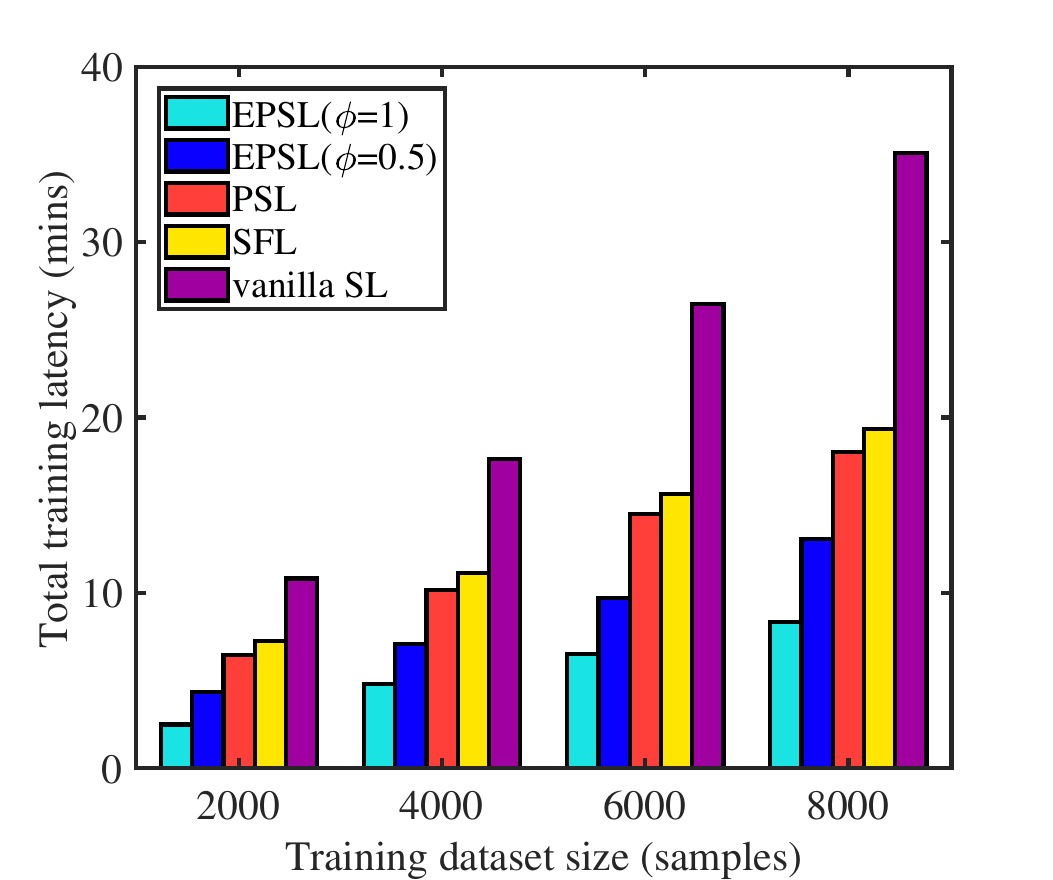}
\vspace{0cm}
\caption{The total training latency for achieving target accuracy on HAM10000 versus the training dataset size under IID setting with $M$=20 and $C$=5.
}
\label{dataset_size}
\end{figure}

Fig.~\ref{dataset_size} shows the total training latency for achieving target accuracy versus the size of training dataset under different SL frameworks. It can be seen that the training latency of vanilla SL, SFL, PSL and EPSL grows as the dataset size increases. The training latency of the vanilla SL is always the highest for different dataset sizes. The reason is that vanilla SL's sequential training scheme results in a significant increase in training latency. The training latency for SFL and PSL is much lower than that for vanilla SL because the parallel training schemes of SFL and PSL are far more effective than sequential training of vanilla SL in accelerating the model training. The EPSL has the lowest total training latency for different dataset sizes. Furthermore,  compared to vanilla SL, SFL and PSL, the training latency for EPSL grows more slowly as the dataset size increases, due to the last-layer gradient aggregation design and the elimination of model exchange. This result demonstrates that the proposed EPSL is better suited for large-scale data model training, which is conducive to widespread implementations of EPSL.

\subsection{Performance Evaluation of the Proposed Resource Management Strategy}

In this subsection, we evaluate the effectiveness of the tailored resource management and layer split strategy in terms of per-round latency. We compare the proposed resource management strategy with four baselines listed below to assess its benefits:

\begin{itemize}
\item \textbf{Baseline a):} The received signal strength (RSS)-based subchannel allocation approach is adopted, which allocates each subchannel to the edge device with the highest RSS in this subchannel. The transmit PSD is set uniformly among edge devices and subchannels while the cut layer is randomly selected.
\item \textbf{Baseline b):} The proposed greedy subchannel allocation approach and power control scheme depicted in Section~\ref{Solution_Approach} are adopted. The cut layer is randomly selected.
\item \textbf{Baseline c):} The RSS-based subchannel allocation approach, the proposed cut layer selection strategy and  power control scheme are adopted.
\item \textbf{Baseline d):} The proposed greedy subchannel allocation approach and cut layer selection strategy are adopted. The transmit PSD is set uniformly among edge devices and subchannels.
\end{itemize}

\begin{figure}[t!]
\centering
\includegraphics[width=6.7cm,height=5.7cm]{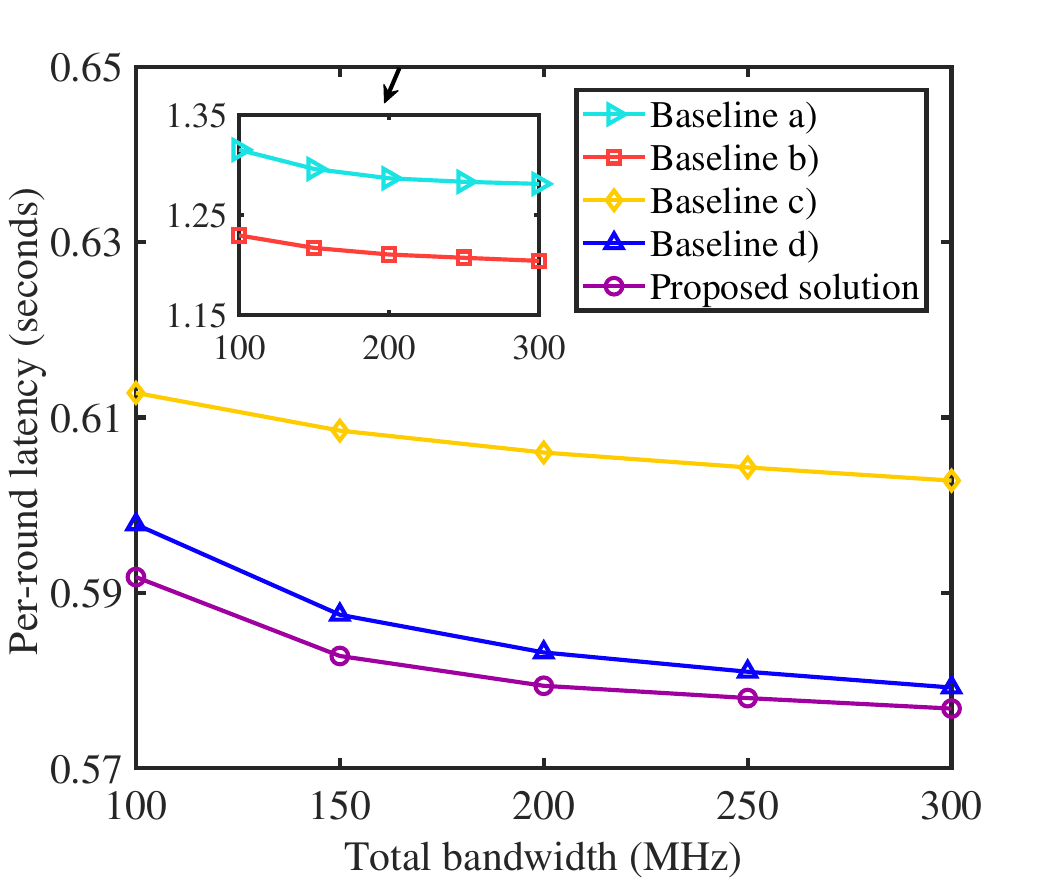}
\vspace{0cm}
\caption{The performance for per-round latency versus the total bandwidth with $\phi = 0.5$, $f_s=5 \times {10^9}$ cycles/s and $d_{max}=200$m.
}
\label{different_subnum_1}
\end{figure}

\begin{figure}[htbp]
\centering
\includegraphics[width=6.7cm,height=5.7cm]{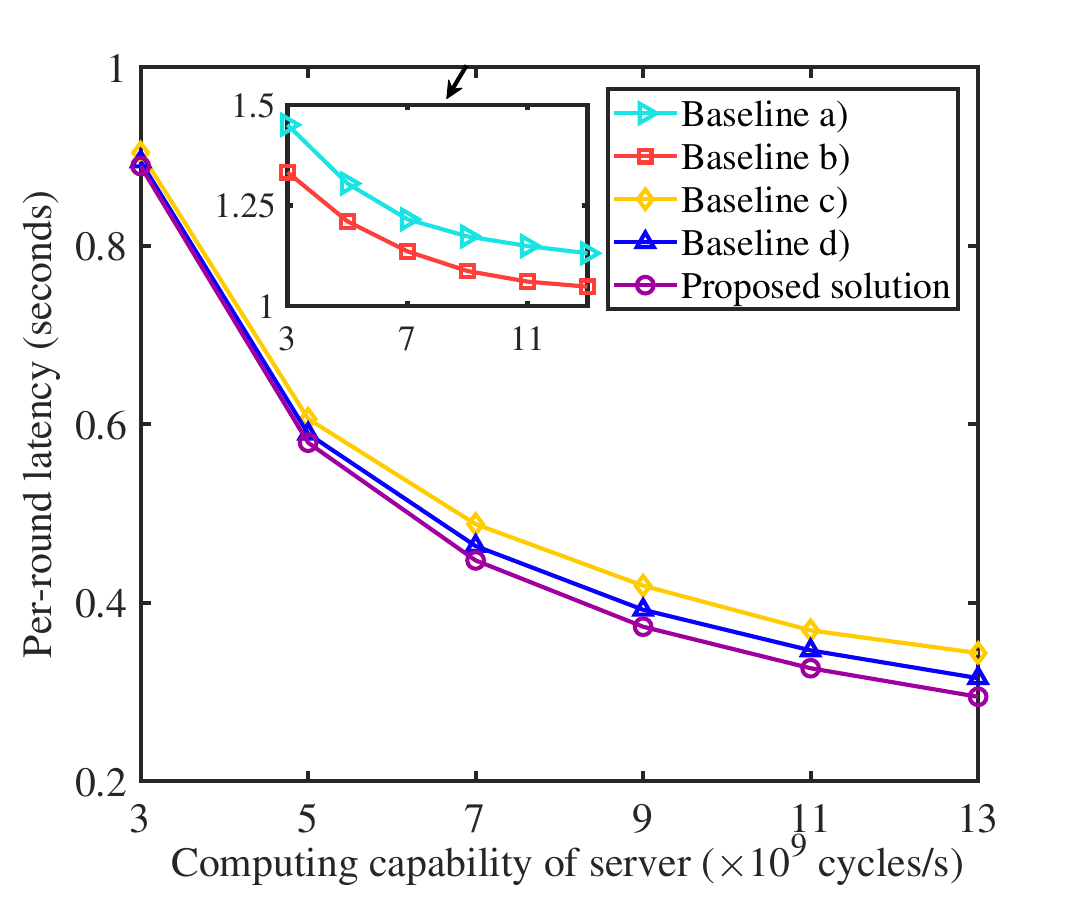}
\vspace{0cm}
\caption{The performance for per-round latency versus the computing capability of server with $\phi = 0.5$, $B_{total}=200$MHz and $d_{max}=200$m.
}
\label{different_server_1}
\end{figure}

Fig.~\ref{different_subnum_1} shows the performance for per-round latency versus the total bandwidth. \rev{The comparison between the proposed solution and the baseline a) reveals that the proposed resource management solution can significantly reduces training latency by approximately 50 \% compared to  counterpart  without optimization.} It is also seen that the training latency decreases as the total bandwidth grows. The reason is that increasing total bandwidth results in higher uplink and downlink transmission rates between edge devices and the server. The reduction in training latency achieved by power control becomes slow as the total bandwidth increases due to the considerable decrease in data exchange latency.  In this case, the per-round latency is primarily determined by the computing latency of the edge devices and the server. The performance gap between baseline c) and proposed solution is widened with larger total bandwidth, due to that the proposed greedy subchannel allocation approach gives priority to reducing the latency of the straggler. \rev{Moreover, we observe that optimizing cut layer selection leads to considerable reduction in training latency, which demonstrates the significance of taking cut layer selection into the optimization for SL. }

Fig.~\ref{different_server_1} illustrates the performance for per-round latency versus the computing capability of the server. It can be observed that the training latency decreases with the increase of the server computing capability. Furthermore, with a more powerful server, the performance improvements brought by power control and subchannel allocation grow. This is because, in this case, the training latency is primarily limited by the computing latency of the edge devices and data exchange latency. Again, we observe that optimizing the cut layer selection brings better performance improvement than power control and subchannel allocation when the server's computing capability varies.

\begin{figure}[t]
\centering
\includegraphics[width=6.7cm,height=5.7cm]{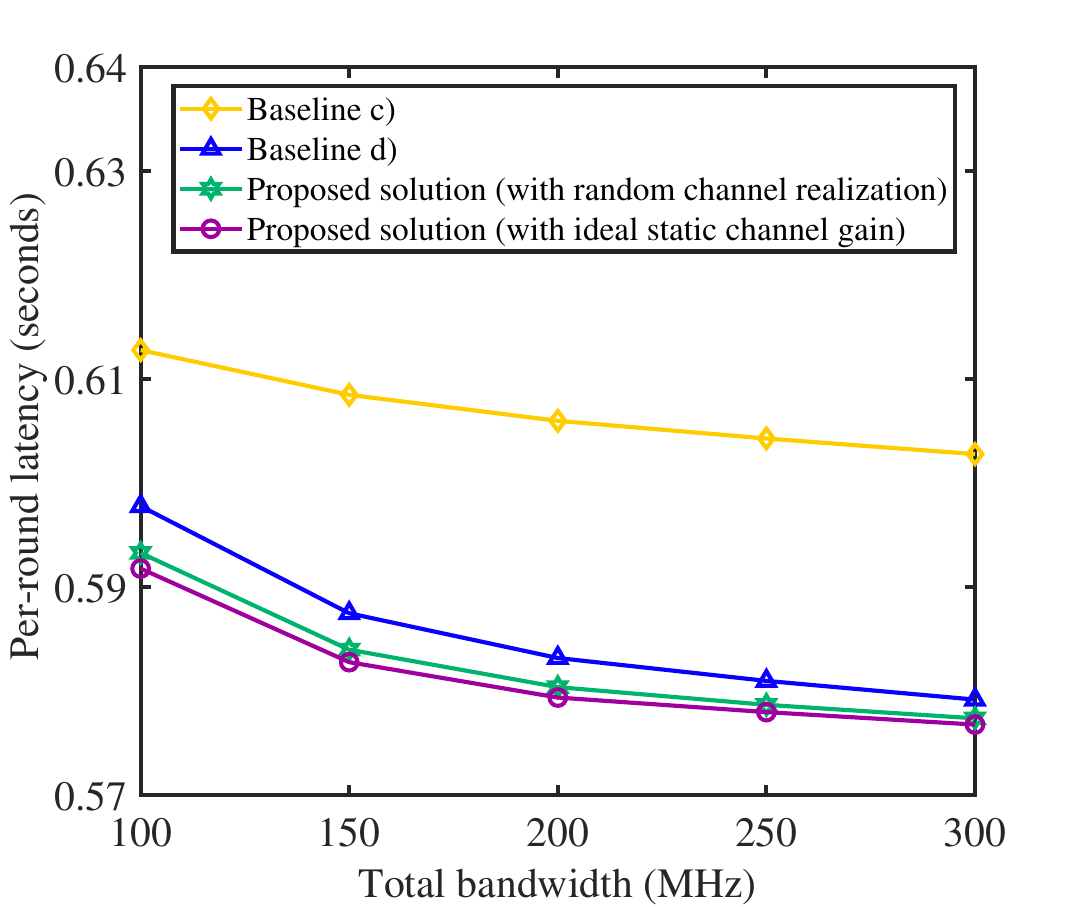}
\vspace{0cm}
\caption{The effect of channel variation on the performance of the proposed solution with $\phi = 0.5$, $B_{total}=200$MHz and $d_{max}=200$m.
}
\label{channel_impact}
\end{figure}

Fig.~\ref{channel_impact}  shows the effect of channel variation on the performance of the proposed solution. Our layer split decision remains unchanged for a period (in the simulations, it remains the same until the model converges), and therefore it is important to evaluate how channel variation would impact its performance. Specifically, we compare the random realization of channel model~\cite{samimi2015probabilistic} in each training round and the ideal static channel channel model (i.e. channel gain remain unchanged, which is unrealistic but used as the benchmark). We observe that channel variation has little impact on the performance of the proposed solution, which demonstrates its robustness in dynamic wireless channel conditions. \rev{This also implies that we do not need to frequently run the BCD-based algorithm unless significant channel variations occur, making the algorithm easy for practical deployment.}

\section{Conclusions}\label{conclusion_section}

In this paper, we have proposed a novel SL framework, efficient parallel split learning (EPSL), to accelerate the model training. EPSL parallelizes the client-side model training, and lowers the dimension of activations' gradients and server's computation workload by performing the last-layer gradient aggregation, leading to significant reduction in training latency. By considering the heterogeneity in channel conditions and computing capabilities at edge devices, we have designed a resource management and layer split strategy to jointly optimize subchannel allocation, power control, and cut layer selection to minimize the training latency for ESPL over the wireless edge networks. While the formulated problem is a mixed-integer nonlinear programming, we decompose it into four less complicated subproblems based on different decision variables, and present an efficient BCD-based algorithm to solve this problem. Simulation results demonstrate that our proposed EPSL framework takes significantly less time budget to achieve a target accuracy than existing benchmarks, and the effectiveness of the tailored resource management and layer split strategy.

This work has demonstrated the potential of integrating EPSL and edge computing paradigm. However, more research attention needs to be paid to addressing label privacy issue (especially for privacy-sensitive data, such as disease diagnosis data) resulting from the label sharing in SL. \rev{Moreover, convergence analysis of EPSL and its generalization to different types of neural network architectures are worth further exploration.}



\ifCLASSOPTIONcaptionsoff
  \newpage
\fi



%


\bibliographystyle{IEEEtran}
\bibliography{NEWmybib}

\end{document}